\documentclass[12pt]{article}
\usepackage{latexsym, amsbsy, amssymb, amsmath, hyperref, epsfig, threeparttable, dcolumn, multirow, booktabs, authblk, pifont, chngcntr, xr, algorithm, algpseudocode, subcaption, tikz, natbib, graphicx, setspace, lscape, longtable, pgfplots, amsthm, xcolor, bm, mathrsfs, enumitem}
\newtheorem{Theorem}{Theorem}
\newtheorem{Corollary}[Theorem]{Corollary}
\newtheorem{Lemma}[Theorem]{Lemma}
\newtheorem{Remark}{Remark}
\newtheorem{Definition}{Definition}
\newtheorem{Assumption}{Assumption}
\newtheorem{Example}{Example}
\newtheorem{Model}{Model}
\newtheorem{Setting}{\normalfont{\textit{Setting}}}

\renewcommand{\theModel}{\arabic{Model}}
\counterwithin{Setting}{Model}
\numberwithin{equation}{section}

\renewcommand{\baselinestretch}{1.2}

\newcommand{\xmark}{\textrm{\ding{55}}}
\newcommand{\newcheckmark}{\textrm{\ding{52}}}

\def\bzero{\boldsymbol{0}}
\def\bone{\boldsymbol{1}}

\def\bxi{\boldsymbol{\xi}}
\def\boldeta{\boldsymbol{\eta}}

\def\beps{\boldsymbol{\epsilon}}

\def\bmu{\boldsymbol{\mu}}

\def\bE{\mathbb{E}}

\def\bR{\mathbb{R}}

\def\upd{\textup{d}}

\def\B{\mathbf{B}}

\def\f{\mathbf{f}}
\def\g{\mathbf{g}}
\def\I{\mathbf{I}}

\def\u{\mathbf{u}}
\def\U{\mathbf{U}}
\def\v{\mathbf{v}}
\def\V{\mathbf{V}}
\def\x{\mathbf{x}}
\def\X{\mathbf{X}}
\def\y{\mathbf{y}}
\def\Y{\mathbf{Y}}
\def\z{\mathbf{z}}
\def\Z{\mathbf{Z}}

\def\calD{\mathcal{D}}
\def\calF{\mathcal{F}}

\def\calI{\mathcal{I}}

\def\calR{\mathcal{R}}

\def\indep{\perp\!\!\!\perp}

\def\dN{\text{Normal}}
\def\dU{\text{Uniform}}

\def\var{\mathop{\textup{var}}\nolimits}
\def\cov{\mathop{\textup{cov}}\nolimits}
\def\dcov{\mathop{\textup{dcov}}\nolimits}
\def\dcorr{\mathop{\textup{dcorr}}\nolimits}

\def\ipcov{\mathop{\textup{ipcov}}\nolimits}
\def\ipcorr{\mathop{\textup{ipcorr}}\nolimits}
\def\supp{\mathop{\textup{supp}}\nolimits}

\DeclareMathOperator*{\argmin}{arg\,min}

\newcommand{\blind}{0}

\begin{document}

\def\spacingset#1{\renewcommand{\baselinestretch}%
{#1}\small\normalsize} \spacingset{1}


\if0\blind
{
  \title{\Large\bf {From Conditional to Unconditional Independence:}\\
Testing Conditional Independence via Transport Maps}
	\author{\small Chenxuan $\rm{He}^{a}$\thanks{Chenxuan He and Yuan Gao are co-first authors.},   Yuan $\rm{Gao}^{b*}$, 
	Liping $\rm{Zhu}^{a}$, 
	and
	Jian $\rm{Huang}^{b,c}$

\vspace{0.2cm}
	{\footnotesize {{$\it^{a}$Institute of Statistics and Big Data, Renmin University of China}}}

	{\footnotesize { {$\it^{b}$Department of Data Science and Artificial Intelligence, The Hong Kong Polytechnic University}}}

{\footnotesize {{$\it^{c}$Department of Applied Mathematics, The Hong Kong Polytechnic University}}}
}
\date{\footnotesize \today}
  \maketitle} \fi

\if1\blind
{
  \bigskip
  \bigskip
  \bigskip
  \begin{center}
    {\Large\bf From Conditional to Unconditional Independence:
Testing Conditional Independence via Transport Maps}
\end{center}
  \medskip
} \fi

\bigskip

\begin{abstract}
Testing conditional independence between two random vectors given a third is a fundamental and challenging problem in statistics, particularly in multivariate nonparametric settings due to the complexity of conditional structures. We propose a novel method for testing conditional independence by transforming it to an unconditional independence test problem.
We achieve this by constructing two transport maps that transform conditional independence into unconditional independence, this substantially simplifies the problem. These transport maps are estimated from data using conditional continuous normalizing flow models. Within this framework, we derive a test statistic and prove its asymptotic validity under both the null and alternative hypotheses. A permutation-based procedure is employed to evaluate the significance of the test. We validate the proposed method through extensive simulations and real-data analysis.  Our numerical studies demonstrate the practical effectiveness of the proposed method for conditional independence testing.
\end{abstract}


\spacingset{1.2} 
\section{Introduction}
\label{sec:intro}

Let $\X \in \bR^{d_\x}$, $\Y \in \bR^{d_\y}$, and $\Z \in \bR^{d_\z}$ be three random vectors.
We are interested in testing whether $\X$ and $\Y$ are conditionally independent given $\Z$ \citep{dawidConditional1979}, formally defined as:
\begin{align}
\label{eq:cond_indep}
H_0:\ \X \indep \Y \mid \Z, \text{ versus } H_1: \X \not \indep \Y \mid \Z.
\end{align}

Conditional independence is a fundamental concept in statistics and machine learning, playing a crucial role in various fields such as causal inference \citep{spirtesCausation1993, pearlCausality2009, spirtesIntroduction2010, constantinouExtended2017} and graphical models \citep{geigerLogical1993, koldanovUniformly2017, fanProjectionbased2020}.
Understanding the conditional independence between random vectors allows researchers to simplify complex models, reduce dimensionality, and enhance interpretability.

Numerous methods have been proposed for testing conditional independence. 
Parametric approaches impose explicit structural assumptions on the conditional distributions $\X \mid \Z$ and $\Y \mid \Z$.
For example, \cite{songTesting2009} adopted a single-index model paired with the Rosenblatt transformation \citep{rosenblattRemarks1952}, while \cite{fanProjectionbased2020} assumed linearity in the conditional distributions.
Though such methods achieve root-$n$ consistency under their parametric constraints, their reliance on strict model specifications renders them vulnerable to misspecification bias, limiting their applicability to complex, non-linear dependencies.

Beyond parametric constraints, kernel-based methods employ kernel functions to measure dependence without parametric assumptions.
For instance, \cite{zhangKernelbased2011} and \cite{huangKernel2022} derived tests using cross-covariance operator in reproducing kernel Hilbert spaces (RKHS).
\cite{wangConditional2015} introduced conditional distance correlation, where they utilized kernel functions to estimate the empirical form.
\cite{caiDistribution2022} formulated conditional independence as mutual independence across transformed variables, estimating dependencies via kernel-based empirical measures.
Though these methods avoid strict parametric assumptions, they either incur substantial computational costs from kernel matrix operations or exhibit reduced statistical power against alternatives as the dimension of $\Z$ grows, even at moderate dimensionality (e.g., dimension of $\Z$ larger than 3).
Recent methodological advances have introduced modern machine learning techniques for conditional independence testing. For example, \cite{senModelpowered2017} formulated conditional independence test as a classification problem, employing gradient-boosted trees and deep neural networks to perform the classification.
\cite{bellotConditional2019}, \cite{shiDouble2021}, \cite{yangConditional2024}, and \cite{duongNormalizing2024} propose utilizing generative model, such as GANs and diffusion models, to learn conditional distributions. They further use these learned distributions to get the residuals. These methods represent modern extensions of the partial correlation concept.

In this work, we propose a framework for conditional independence testing based on transport maps.
At the population level, we demonstrate that two invertible maps from the conditional distributions to simple reference distributions allow us to formulate conditional independence test as unconditional independence test, thereby simplifying the formulation. We employ conditional continuous normalizing flows and independence measures to effectively address conditional independence testing.
Our contributions are summarized as follows:

\begin{enumerate}
\item \textit{A novel equivalent formulation of conditional independence.}
We demonstrate that testing conditional independence between $\X$ and $\Y$ given $\Z$ can be simplified to testing unconditional independence through invertible transformations.
By transforming $(\X, \Z)$ and $(\Y, \Z)$ into new variables (e.g., Gaussian variables) independent of $\Z$, conditional independence becomes equivalent to unconditional independence.
We rigorously prove this equivalence in Lemma~\ref{lemma:independence}.
This formulation significantly reduces the complexity associated with conditional independence structure.

\item \textit{A model-free test statistic.}
We introduce a model-free test statistic for conditional independence testing. By using the continuous normalizing flow as a transport map, we can apply popular independence measures, such as distance correlation, which do not depend on specific model assumptions about the data distribution.
The asymptotic validity is given in Theorem~\ref{thm:asymptotics}.

\item \textit{Performance across dimensions.} Through extensive simulations, we demonstrate that the proposed test maintains robust performance across various dimensions for variables $\X$, $\Y$, and $\Z$, with each dimension ranged from 1 to 50.

\item \textit{Robustness to independence measures.} After transforming a conditional independence problem into an unconditional independence one, our test remains robust to the selection of dependence measures. We illustrate this by comparing the results of the proposed method using distance correlation \citep{szekelyMeasuring2007} and improved projection correlation \citep{zhangProjective2024} in
     Supplementary Materials \ref{appendix:sec:compare_DC_IPC}  .
\end{enumerate}

We present an application of the proposed method to the problem of sufficient dimension reduction, specifically in testing whether an empirically estimated data representation satisfies the sufficiency requirement, as discussed in Section~\ref{sec:sim}.

Our approach utilizes a permutation-based method for computing $p$-values. This method is supported by both simulation studies and real-data analysis, which collectively demonstrate its robustness, empirical validity, and practical utility. Theoretical justification for the permutation-based method is provided in Theorem~\ref{thm:permutation_p}, with numerical evaluations detailed in Section~\ref{sec:sim}.

The remainder of the paper is organized as follows: We provide some preliminaries in Section~\ref{sec:preliminaries} and introduce the methodology in Section~\ref{sec:methodology}. Simulation studies and an application to dimension reduction are presented in Sections~\ref{sec:sim} and~\ref{sec:real_data}. A brief conclusion is provided in Section~\ref{sec:conclusion}.


\emph{Notation.}
Throughout this paper, uppercase letters such as $X$, $Y$, and $Z$ denote random variables, with their realizations represented by the corresponding lowercase letters $x$, $y$, and $z$. Bold uppercase letters such as $\X$, $\Y$, and $\Z$ denote random vectors, while bold lowercase letters such as $\x$, $\y$, and $\z$ denote their realizations as real-valued vectors. The Euclidean norm of a vector $\x \in \mathbb{R}^d$ is denoted by $\|\x\|$.
The Gamma function is denoted by $\Gamma(\cdot)$, the $d$-dimensional identity matrix by $\I_d$, and the indicator function by $\bone(\cdot)$. We use $\stackrel{d}{=}$ to denote equal in distribution and $\stackrel{p}{\to}$ to denote convergence in probability. The support of a random vector is denoted by $\supp(\cdot)$.
For two sequences \(a_n\) and \(b_n\), the notation \(a_n = O(b_n)\) or \(a_n \lesssim b_n\) indicates that there exists a constant \(C_0\) such that \(|a_n| \le C_0 |b_n|\) as $n$ goes to infinity.

\section{Basic Formulation}
\label{sec:preliminaries}

In this section, we present a basic formulation of our proposed method. The core idea is to employ transport maps to transform a conditional independence test problem into an unconditional one. This transformation is detailed in Section~\ref{sec:ci2independence}. We then discuss the implementation of transport maps using continuous normalizing flow models in Section~\ref{sec:ccnf}. Following the application of the transport maps, we introduce independence measures in Section~\ref{sec:dc}.

\subsection{From Conditional Independence to Independence}
\label{sec:ci2independence}
To begin with, we establish an equivalence between conditional independence and unconditional independence by invertible transport maps.
Recall the formulation for conditional independence testing given in \eqref{eq:cond_indep}.
Denote the original jointly distribution triplet as $(\X, \Y, \Z) \in \bR^{d_\x+d_\y+d_\z}$.
Suppose there are two invertible transport maps that push the conditional distributions $\X \mid \Z$ and $\Y \mid \Z$ to two simple reference distributions (e.g., standard normal distributions).
We say a function $f(\x, \z): \bR^{d_\x} \times \bR^{d_\z} \to \bR^{d_\x}$ is invertible with respect to the first argument if, for every fixed $\z \in \bR^{d_\z}$, there exists an inverse function $f^{-1}: \bR^{d_\x} \times \bR^{d_\z} \to \bR^{d_\x}$ such that $f^{-1} (f(\x, \z), \z) = f (f^{-1}(\x, \z), \z) = \x$ for all $\x \in \bR^{d_\x}$ and $\z \in \bR^{d_\z}$.
We have the following Lemma~\ref{lemma:independence}.
\begin{Lemma}[Conditional independence to unconditional independence]
\label{lemma:independence}
For the jointly distributed triplet $(\X, \Y, \Z) \in \bR^{d_\x + d_\y + d_\z}$, suppose there exist two invertible transport maps (with respect to their first arguments) $\bxi = \f(\X, \Z)$ and $\boldeta = \g(\Y, \Z)$ such that:
\begin{align*}
\bxi = \f(\X, \Z) \stackrel{d}{=} & \big[ \f(\X, \Z) \mid \Z = \z \big], \quad \forall \z \in \supp(\Z),\\
\boldeta = \g(\Y, \Z) \stackrel{d}{=} & \big[ \g(\Y, \Z) \mid \Z = \z \big], \quad \forall \z \in \supp(\Z).
\end{align*}
Then, it holds that
$\bxi \indep \Z, \quad \boldeta \indep \Z.$
Furthermore, we have: 
\begin{align*}
	\X \indep \Y \mid \Z \iff \boldeta \indep (\bxi, \Z) \iff \bxi \indep (\boldeta, \Z).
\end{align*}
\end{Lemma}

\begin{Remark}
	Lemma~\ref{lemma:independence} establishes equivalences between conditional and unconditional independence via transport maps.
	This result aligns with, yet generalizes, earlier frameworks in the literature.
	For instance, let $F_1(\X \mid \Z)$ and $F_2(\Y \mid \Z)$ denote the conditional distributions of $\X$ and $\Y$ given $\Z$, and let $F_3(\Z)$ represent the marginal distribution of $\Z$. Under the assumption that $(F_1(\X \mid \Z), F_2(\Y \mid \Z)) \indep \Z$, \cite{zhouTest2020} demonstrated that conditional independence reduces to the unconditional independence $F_1(\X \mid \Z) \indep F_2(\Y \mid \Z)$. Similarly, \cite{caiDistribution2022} characterized conditional independence via the mutual independence of $F_1(\X \mid \Z)$, $F_2(\Y \mid \Z)$, and $F_3(\Z)$.
	While these results provide foundational insights, they are inherently constrained to specific distributional representations.
	Our work extends this paradigm by utilizing transport maps to create a more flexible equivalence framework, and avoid the assumption of joint independence in formulating the problem.
\end{Remark}

By applying Lemma~\ref{lemma:independence}, we can convert the conditional independence test into an unconditional independence test.
This transformation simplifies the analysis and allows us to leverage classical statistical measures of independence.
Detailed proofs of Lemma~\ref{lemma:independence} can be found in the Supplementary Materials
\ref{sec:proof:lemma:independence}.

Lemma~\ref{lemma:independence} establishes a practical framework for conducting conditional independence testing. The conditional Continuous Normalizing Flow model (CNF, \citealt{liuFlow2023}, \citealt{albergoBuilding2022}, \citealt{lipmanFlow2022}), a state-of-the-art generative model, is able to formulate the transport map and is proven to be well-posed under certain regularity conditions \citep{gaoGaussian2024}. Building on this, the next section presents an implementation of Lemma~\ref{lemma:independence} using conditional CNFs.

\subsection{Conditional Continuous Normalizing Flows}
\label{sec:ccnf}

In this section, we introduce the conditional CNFs with flow matching \citep{liuFlow2023, albergoBuilding2022, lipmanFlow2022}.
Let $\X_0 \sim \dN(\bzero, \I_{d_\x})$, $\Y_0 \sim \dN(\bzero, \I_{d_\y})$ be two random vectors independent of $(\X, \Y, \Z)$. We illustrate the conditional CNF for the pair of distributions, the distribution of $\X_0$ and the conditional distribution of $(\X \mid \Z)$; and similarly for $\Y_0$ and $(\Y \mid \Z)$.
We introduce a specific type of CNFs, namely, the rectified flow \citep{liuRectified2022, liuFlow2023}.
A rectified flow is constructed based on a linear stochastic interpolation between $\X_0$ and $\X$ over time $t \in[0,1]$:
\begin{align}
\label{eq:rect_flow}
\X_t = (1-t) \X_0 + t \X, \quad \forall t \in [0,1].
\end{align}
When $t=0$, since $\X_0 \indep (\X, \Z)$, it follows that $(\X_0\mid \Z) = \X_0$ in distribution. When $t=1$, we have $(\X \mid \Z) = (\X_1 \mid \Z)$.
The process defined in \eqref{eq:rect_flow} indicates that as $t$ goes from 0 to 1, the conditional distribution of $(\X_t \mid \Z)$ progressively evolves from the standard normal distribution to the
conditional distribution of $(\X_1 \mid \Z), t \in [0,1].$

We now present the definition of the velocity field related to the conditional distribution of
$(\X_t \mid \Z)$.

\begin{Definition}[Velocity field]
\label{def:velocity_field}
For the time-differentiable random process $\{(\X_t \mid \Z), t\in [0,1]\}$ defined in \eqref{eq:rect_flow}, its velocity field $\v_\x(t, \x, \z)$ is defined as
\begin{align*}
\v_\x(t, \x, \z) =& \bE \left\{ \frac{\upd \X_t}{\upd t} \ \bigg| \ \X_t=\x, \Z=\z \right\} \\
=& \bE \left( \X - \X_0 \mid \X_t=\x, \Z=\z \right), \quad \forall \x \in \supp(\X_t), \z \in \supp(\Z).
\end{align*}
\end{Definition}

By integrating the velocity field starting from $\X_0$, we derive the forward process:
\begin{align}
\label{eq:X_t_0-1}
(\X_t^\sharp \mid \Z) = \X_0 + \int_0^t \v_\x(s, \X_s^\sharp, \Z) \upd s, \quad \forall t \in [0,1], \quad \X^\sharp_0=\X_0,
\end{align}
where we use the notation $(\X_t^\sharp \mid \Z)$ to represent the state of the process at time $t$ conditional on $\Z$. We will use this notation repeatedly below.
Conversely, by integrating backward from 
$\X_1=\X,$ we obtain the reverse process:
\begin{align}
\label{eq:X_t_1-0}
(\X_t^\dagger \mid \Z) = \X_1 + \int_1^t \v_\x(s, \X_s^\dagger, \Z) \upd s, \quad \forall t \in [0,1], \quad (\X^\dagger_1 \mid \Z) = (\X \mid \Z).
\end{align}

The forward process $(\X_t^\sharp \mid \Z)$ and the reverse process $(\X_t^\dagger \mid \Z)$ have been shown to possess the marginal preserving property, as detailed in Lemma~\ref{lemma:marginal_preserving} below.

\begin{Lemma}[Marginal preserving property]
\label{lemma:marginal_preserving}
Under several regularity conditions, these two processes $(\X_t^\sharp \mid \Z)$ and $(\X_t^\dagger \mid \Z)$ defined in \eqref{eq:X_t_0-1} and \eqref{eq:X_t_1-0} are well-posed and unique.
In addition, the following property holds:
\begin{align*}
(\X_t \mid \Z = \z) \stackrel{d}{=} (\X_t^\sharp \mid \Z = \z) \stackrel{d}{=} (\X_t^\dagger \mid \Z = \z), \quad \text{for any } (\z, t) \in \supp(\Z) \times [0,1],
\end{align*}
where $(\X_t \mid \Z)$ is defined in \eqref{eq:rect_flow}.
\end{Lemma}

The marginal preserving property is well-established in the literature.
For instance, Theorem~5.1 and Corollary~5.2 in \cite{gaoGaussian2024} establish this property under suitable assumptions. Similar results can also be found in \cite{liuRectified2022}, \cite{albergoStochastic2023}, and \cite{huangConditional2024}.

According to Lemma~\ref{lemma:marginal_preserving}, the velocity field $\v_\x(t, \x, \z)$ enables the transformation of the distribution $\X_0$ into $(\X \mid \Z)$, and vice versa.
Furthermore, since the ordinary differential equation (ODE) in \eqref{eq:rect_flow} and the integration process defined in \eqref{eq:X_t_1-0} are deterministic, the transformation is one-to-one and invertible. This property is not shared by the stochastic differential equation-based methods, such as diffusion models \citep{hoDenoising2020, songDenoising2020}.
Consequently, the transport map induced by the rectified flow is invertible and meets the conditions outlined in Lemma~\ref{lemma:independence}.
Thus, we introduce two rectified flows:
\begin{align*}
\X_t  =& (1-t) \X_0 + t \X , \quad \forall t \in [0,1],\\
\Y_t =& (1-t) \Y_0 + t \Y, \quad \forall t \in [0,1].
\end{align*}

These two processes induce two velocity fields $\v_\x(t, \x, \z)$ and $\v_\y(t, \y, \z)$ as given in Definition~\ref{def:velocity_field}. At the population level, by integrating the velocity fields starting from $\X $ and $\Y,$ we obtain:
\begin{align*}
(\X_0^\dagger \mid \Z) &= \X + \int_1^0 \v_\x(s, \X_s^\dagger, \Z) \upd s, \quad (\X_1^\dagger \mid \Z) = (\X \mid \Z)\\
(\Y_0^\dagger \mid \Z) &= \Y  + \int_1^0 \v_\y(s, \Y_s^\dagger, \Z) \upd s, \quad (\Y_1^\dagger \mid \Z) = (\Y \mid \Z).
\end{align*}

By Lemma~\ref{lemma:marginal_preserving}, we have $(\X_0^\dagger \mid \Z) \stackrel{d}{=} \X_0$ and $(\Y_0^\dagger \mid \Z) \stackrel{d}{=} \Y_0$.
We define:
\begin{align}
\label{eq:f_x}
\bxi :=& (\X_0^\dagger \mid \Z) \stackrel{d}{=} \X_0, \\
\label{eq:f_y}
\boldeta :=& (\Y_0^\dagger \mid \Z) \stackrel{d}{=} \Y_0,
\end{align}
where $\bxi$ is a function of $\X$ and $\Z$, and $\boldeta$ is a function of $\Y$ and $\Z$.
The pair $(\bxi, \boldeta)$ is exactly what we expect in Lemma~\ref{lemma:independence}. We illustrate this idea in Example~\ref{eg:simple_ode} below.

\begin{Example}
\label{eg:simple_ode}
We illustrate Lemmas~\ref{lemma:independence} and \ref{lemma:marginal_preserving} through a Gaussian example.
Let $Z \sim \dN(0, 1)$, with $X\mid Z \sim \dN(Z, 1)$ and $Y\mid Z \sim \dN(Z, 1)$.
By construction, $X \indep Y \mid Z$.
The linear interpolation between $X_0$ and $X$ is
$X_t  = (1-t) X_0 + t X, \quad \forall t \in [0, 1].$
Conditional on $Z$, the joint distribution of $(X, X_0, X_t)^\top$ is $\dN\left(\bmu, \Sigma\right)$, where $\bmu = (Z, 0, Z-tZ)^\top$ and
\begin{align*}
\Sigma = \left(\begin{array}{ccc}
    1 & 0 & t \\
    0 & 1 & 1-t \\
    t & 1-t & 1-2t-t^2
\end{array}\right).
\end{align*}
From the properties of joint normal distributions, the velocity field is given by
\begin{align*}
v_x(t, x, z) = \bE(X- X_0\mid X_t=x, Z=z) = \frac{2t-1}{1-2t-2t^2} \{x - (1-t) z\} = \frac{\upd x}{\upd t}.
\end{align*}

Solving this ODE backward in time with terminal condition $X_1 = x$, we obtain that $X_0^\dagger=z-x$.
Consequently, we have $\xi = Z- X \sim \dN(0, 1)$.
Analogously, $\eta = Z- Y \sim \dN(0, 1)$.
Thus, $\xi \indep (\eta, Z)$ and $\eta \indep (\xi, Z)$ are satisfied.
\end{Example}

\subsection{Independence Measures}
\label{sec:dc}

Based on Lemma~\ref{lemma:independence}, after transforming the conditional variables $(\X \mid \Z)$ and $(\Y \mid \Z)$ into $\bxi$ and $\boldeta$, respectively, the conditional independence test problem $\X \indep \Y \mid \Z$ is simplified to testing $\bxi \indep (\boldeta, \Z)$ or $\boldeta \indep (\bxi, \Z)$. 

Several popular independence measures are suitable for use, such as Distance Correlation (DC, \citealt{szekelyMeasuring2007}), Projection Correlation (PC, \citealt{zhuProjection2017}), and Improved Projection Correlation (IPC, \citealt{zhangProjective2024}).
They are all equal to zero if and only if two random vectors are independent.

Given the computational efficiency of DC and IPC, we recommend using these two measures to assess the independence.
Our numerical studies suggest that the test results remain robust regardless of the chosen dependence measure. For further details, please refer to Appendix \ref{appendix:sec:compare_DC_IPC}.

In the following, we illustrate the idea of independence measures by distance covariance and distance correlation. An introduction for IPC is given in Appendix~\ref{sec:ipc}.

Let $\U$ and $\V$ be two arbitary random vectors. The distance covariance between them  is defined
as \citep{szekelyMeasuring2007}:
\begin{align}
\label{eq:dc_pop}
\dcov^2(\U, \V)
=& \int \left\| f_{\U, \V}(\u, \v) - f_{\U}(\u) f_{\V}(\v) \right \|^2  w(\u, \v) \bE_\u \upd \v,
\end{align}
where $f_{\U, \V}(\cdot, \cdot)$ is the joint characteristic function of $(\U, \V)$, $f_{\U}(\cdot)$ and $f_{\V}(\cdot)$ are the marginal characteristic functions of $\U$ and $\V$, and $w(\cdot, \cdot)$ is the weight function defined by
$
w(\u, \v) = \left\{ c_{d_\x} c_{d_\y} \|\u\|^{1+d_\x} \|\v\|^{1+d_\y} \right\}^{-1},
$
with $c_d = \pi^{(1+d)/2}/\Gamma((1+d)/2)$.

Since $\U \indep \V \iff f_{\U, \V}(\u, \v) = f_{\U}(\u) f_{\V}(\v)$, the distance covariance is nonnegative and equals to zero if and only if $\U$ and $\V$ are independent.
Let $(\U_1, \V_1)$, $(\U_2, \V_2)$, and $(\U_3, \V_3)$ be independent copies of $(\U, \V)$.
If $\bE\|\U\|^2 < \infty$ and $\bE \|\V\|^2 < \infty$, the distance covariance defined in \eqref{eq:dc_pop} can be expressed as \citep{szekelyMeasuring2007}
\begin{align}
\nonumber
\dcov^2(\U, \V) =& \bE\left( \|\U_1 - \U_2\| \| \V_1 - \V_2 \| \right) + \bE\|\U_1 - \U_2\| \bE \|\V_1-\V_2\| \\
& - 2 \bE \left(\|\U_1 - \U_2\| \|\V_1-\V_3\|\right). \label{eq:dc_pop_simplified}
\end{align}
The distance correlation is give by
\begin{align*}
\dcorr^2(\U, \V) = \frac{\dcov^2(\U, \V)}{\dcov(\U, \U) \dcov(\V, \V)}.
\end{align*}

With the ability to transform conditional independence into an unconditional independence framework, along with tools to measure unconditional independence, we are now prepared to introduce the methodology in Section~\ref{sec:methodology}.

\section{Methdology}
\label{sec:methodology}

In this section, we begin by introducing our test statistic in Section~\ref{sec:test_statistic}. Our method employs a sample splitting approach, and the asymptotic properties of the test statistic are discussed in Section~\ref{sec:test_statistic_asymptotics} under appropriate splitting conditions. To enhance the power performance and address the limitations of splitting, we propose using multiple splitting in Section~\ref{sec:multiple_splitting}.

\subsection{Test Statistic}
\label{sec:test_statistic}

For the triplet $(\X, \Y, \Z)$, we transforme the conditional distributions $(\X \mid \Z)$ and $(\Y \mid \Z)$ into $\bxi$ and $\boldeta$ by conditional CNFs.
The dependence between the transformed variables is then assessed via independence measures.
We develop the empirical counterpart of this procedure to perform the conditional independence test.

Consider the observed dataset $\calD_n = \{(\X_i, \Y_i, \Z_i)\}_{i=1}^n$.
We randomly split the dataset into two disjoint subsets, denoted by two index sets $\calI_1$ and $\calI_2$, where $|\calI_1| = n_1$, $|\calI_2| = n_2$, $\calI_1 \cap \calI_2 = \emptyset$, and $\calI_1 \cup \calI_2 = \{ 1, \ldots, n \}$.
Denote the training dataset as $\calD_{n_1} = \{(\X_i, \Y_i, \Z_i), i \in \calI_1\}$ and the test dataset as $\calD_{n_2} = \{(\X_i, \Y_i, \Z_i), i \in \calI_2\}$.

The proposed test statistic is constructed through four steps.
First, we estimate velocity fields via deep neural networks based on training dataset $\calD_{n_1}$.
Next, we compute the transformed variables $\bxi$ and $\boldeta$ using the maps defined in~\eqref{eq:f_x} and~\eqref{eq:f_y}.
Third, we construct the test statistic $T_n$ as the sample version of the chosen independence measure based on $\calD_{n_2}$.
Finally, we compute the $p$-value using permutation-based methods.
We now give detailed explanations of the four steps.

\textit{Step 1. Estimation of the velocity fields $\v_\x(t, \x, \z)$ and $\v_\y(t, \y, \z)$.}

In the first step, we estimate the velocity fields using deep neural networks.
As established in Definition~\ref{def:velocity_field}, these velocity fields correspond to conditional expectations, making the least squares estimator a natural choice for their estimation.
Operationally, we augment the observed dataset $\calD_{n_1}$ with three auxiliary random samples: $\{\X_{0,i}, i \in \calI_1\} \sim \dN(\bzero, \I_{d_\x})$, $\{\Y_{0,i}, i \in \calI_1 \} \sim \dN(\bzero, \I_{d_\y})$, and $\{t_i, i \in \calI_1 \} \sim \dU([0,1])$.
The velocity field estimators are obtained by minimizing the empirical squared losses:
\begin{align}
\label{eq:hat_vx}\hat \v_\x(t, \x, \z) &= \argmin_{\v_\x \in \calF_{\x}} \frac{1}{n_1} \sum_{i \in \calI_1} \|\v_\x(t_i, \X_i, \Z_i) - (\X_i - \X_{0, i})\|_2^2, \\
\label{eq:hat_vy}\hat \v_\y(t, \y, \z) &= \argmin_{\v_\y \in \calF_{\y}} \frac{1}{n_1} \sum_{i \in \calI_1} \|\v_\y(t_i, \Y_i, \Z_i) - (\Y_i - \Y_{0, i})\|_2^2,
\end{align}
where $\calF_{\x}$ and $\calF_{\y}$ refer to neural network function classes.
These estimated velocity fields then serve as the foundation for computing the transformed variables $\bxi$ and $\boldeta$ in the subsequent step.

\textit{Step 2. Estimation of $\bxi$ and $\boldeta$.}

Given the estimated velocity fields, the estimted transport map is denoted as
\begin{align}
	\label{eq:hat_bxi} (\hat \X_t^\dagger \mid \Z) &= \X + \int_1^t \hat \v_\x(s, \hat \X_s^\dagger, \Z) \upd s, \quad \forall t \in [0, 1], \quad (\hat \X_1^\dagger \mid \Z) = (\X \mid \Z) \\
	\label{eq:hat_boldeta} (\hat \Y_t^\dagger \mid \Z) &= \Y + \int_1^t \hat \v_\y(s, \hat \Y_s^\dagger, \Z) \upd s, \quad \forall t \in [0, 1], \quad (\hat \Y_1^\dagger \mid \Z) = (\Y \mid \Z).
\end{align}
Similar to $\bxi=(\X_0^\dagger \mid \Z)$ and $\boldeta = (\Y_0^\dagger \mid \Z)$ given in \eqref{eq:f_x}--\eqref{eq:f_y}, we define $\hat \bxi = (\hat \X_0^\dagger \mid \Z)$ and  $\hat \boldeta = (\hat \Y_0^\dagger \mid \Z)$.

We augment the original dataset to $\calD_{n_2} = \{(\X_i, \Y_i, \Z_i, \bxi_i, \boldeta_i), i \in \calI_2\}$, where $\bxi$ and $\boldeta$ are unobservable and need to be estimated.
We denote the estimated version as $\calD_{n_2}^{\prime} = \{(\X_i, \Y_i, \Z_i, \hat{\bxi}_i, \hat{\boldeta}_i), i \in \calI_2\}$.
After obtaining $\calD_{n_2}^{\prime}$, we proceed to the next step to construct the test statistic.

\textit{Step 3. Construction of the test statistic by $\calD_{n_2}^\prime$.}

We compute the empirical version of the chosen independence measure based on $\calD_{n_2}^{\prime} = \{(\X_i, \Y_i, \Z_i, \hat{\bxi}_i, \hat{\boldeta}_i), i \in \calI_2\}$ to construct the test statistic. We demonstrate this procedure using distance correlation.

Following Lemma~\ref{lemma:independence}, we can test one of these two independence hypotheses: $\boldeta \indep (\bxi, \Z)$ or $\bxi \indep (\boldeta, \Z)$. 
Let $\U$ and $\V$ represent the paired vectors for testing.
For example, to test the independence $\boldeta \indep (\bxi, \Z)$, we define $\hat \U = \hat \boldeta$ and $\hat \V=(\hat \bxi^\top, \Z^\top)^\top$ using the estimated transport maps.
Let $\U = \boldeta$ and $\V=(\bxi^\top, \Z^\top)^\top$ denote the  corresponding true (but unobserved) counterpart.

The sample distance covariance is computed as:
\[\dcov_{n_2}^2(\hat \U, \hat \V) = S_1 + S_2 - 2 S_3,
\]
where
\begin{align*}
S_1 =& \frac{1}{n_2^2} \sum_{k,l \in \calI_2} \| \hat \U_k - \hat \U_l \| \| \hat \V_k - \hat \V_l \|, \\
S_2 =& \frac{1}{n_2^2} \sum_{k,l \in \calI_2} \| \hat \U_k - \hat \U_l \| \frac{1}{n_2^2} \sum_{k,l \in \calI_2} \| \hat \V_k - \hat \V_l \|, \\
S_3 =& \frac{1}{n_2^3} \sum_{k \in \calI_2} \sum_{l,m \in \calI_2} \| \hat \U_k - \hat \U_l \| \| \hat \V_k - \hat \V_m \|.
\end{align*}

The test statistic is defined as the squared distance correlation:
\begin{align}
\label{eq:T_n}
T_{n_2} = \dcorr_{n_2}^2(\hat \U, \hat \V) = \frac{\dcov_{n_2}^2 (\hat \U, \hat \V)}{\dcov_{n_2} (\hat \U, \hat \U) \dcov_{n_2} (\hat \V, \hat \V)},
\end{align}
where $\dcov_{n_2} (\hat \U, \hat \U)$ and $\dcov_{n_2} (\hat \V, \hat \V)$ are computed analogously to $\dcov_{n_2} (\hat \U, \hat \V)$.
We then compute the $p$-value by the permutation-based method as the following step.

\textit{Step 4. Obtaining the $p$-value by permutation.}

To obtain the $p$-value for $T_{n_2}$, we first randomly permute $\{(\hat \U_i, \hat \V_i), i \in \calI_2\}$ as $\{(\hat \U_i, \hat \V_{\pi_b(i)}), i \in \calI_2\}$, where $\pi_b$ is a uniformly random permutation of $\calI_2$.
A permuted statistic $T_{n_2, b}$ is then computed using \eqref{eq:T_n} on $\{(\hat \U_i, \hat \V_{\pi_b(i)}), i \in \calI_2\}$.
We replicate the permutation $B$ times and calculate:
\begin{align}\label{eq:permuted_p}
p_B = \frac{1}{B} \sum_{b=1}^B \bone(T_{n_2, b} \ge T_{n_2}).
\end{align}

We summarize the full procedure in Algorithm~\ref{alg:permutation_test} below.
\begin{algorithm}[H]
\caption{FlowCIT: Conditional independence test based on CNFs.}
\label{alg:permutation_test}
\begin{algorithmic}[1]
	\Require Dataset $\calD_n= \{(\X_i, \Y_i, \Z_i)\}_{i=1}^n$; Randomly split index sets $\calI_1$ and $\calI_2$; Number of permutations $B$.
	\Ensure Test statistic $T_{n_2}$; $p$-value.
	\State Generate $\{\X_{0,i}, i \in \calI_1\} \sim \dN(\bzero, \I_{d_\x})$, $\{\Y_{0,i}, i \in \calI_1 \} \sim \dN(\bzero, \I_{d_\y})$, and $\{t_i, i \in \calI_1 \} \sim \dU([0,1])$.
	\State (\textit{Step 1})  Estimate the velocity fields $\v_\x$ and $\v_\y$ by $\{(\X_i, \Y_i, \Z_i), i \in \calI_1\}$, $\{\X_{0,i}, i \in \calI_1\}$, $\{\Y_{0,i}, i \in \calI_1\}$, $\{t_i, i \in \calI_1\}$ as given in \eqref{eq:hat_vx} and \eqref{eq:hat_vy}.
	\State (\textit{Step 2}) Compute $\hat \bxi$ and $\hat \boldeta$ based on $\hat \v_\x$ and $\hat \v_\y$ as given in \eqref{eq:hat_bxi} and \eqref{eq:hat_boldeta}.
	\State (\textit{Step 3})  Compute the test statistic $T_{n_2}$ by an independence measure based on the test dataset (e.g., distance correlation by \eqref{eq:T_n}).
	\State (\textit{Step 4})
	\textbf{For {$b = 1, \dots, B$} do}
	\begin{itemize}
		\item Generate a uniformly random permutation $\pi_b$ of $\{1,2,\dots,n\}$.
		\item Denote the permuted dataset as $\calD_{n_2, b}^{\prime}$.
		\item Compute the permuted test statistic $T_{n_2, b}$ based on $\calD_{n_2, b}^{\prime}$.
		\item \textbf{End for}
	\end{itemize}
	\State Calculate the $p$-value as given in \eqref{eq:permuted_p}.
	\State \Return $T_{n_2}$; $p$-value.
\end{algorithmic}
\end{algorithm}

\subsection{Asmyptotics of the Test Statistic}
\label{sec:test_statistic_asymptotics}

As outlined in Algorithm~\ref{alg:permutation_test}, our procedure involves a four-step estimation process: estimating the velocity field, estimating the transformed pairs, calculating the dependence measure of the transformed pairs, and getting the $p$-value.

In this section, we provide a theoretical analysis of the proposed test.
We derive the asymptotic distribution of the test statistic and discuss how to properly split the data in Section~\ref{sec:test_statistic_asymptotics}.
First, we formalize the convergence requirements for the estimated velocity fields $\hat \v_\x$ and $\hat \v_\y$.

\begin{Assumption}
	\label{con:converge_velocity_field}
	As $n \to \infty$, the $L_2$-errors of the flow matching estimators satisfy:
	\begin{align*}
	\bE_{(t, \X, \Z)}
	\left\|\hat \v_\x(t, \X, \Z) - \v_\x(t, \X, \Z) \right\|^2 &= O_p(n_1^{- 2 \kappa_1}), \\
	\bE_{(t, \Y, \Z)}
	\left\|\hat \v_\y(t, \Y, \Z) - \v_\y(t, \Y, \Z) \right\|^2 &= O_p(n_1^{- 2 \kappa_2}).
	\end{align*}
\end{Assumption}

The convergence results stated in Assumption~\ref{con:converge_velocity_field} align with recent theoretical advances in flow matching \citep{gaoConvergence2024, fukumizuFlow2025, caiMinimax2025, huangConditional2024}.
The parameters $\kappa_1$ and $\kappa_2$ are rates for nonparametric regression, typically ranges in $(0, 1/2)$.

Based on Assumption~\ref{con:converge_velocity_field}, it can inferred that the estimated transport map satisfy,
\begin{align*}
	\bE_{(\X, \Z)} \big\| \hat \bxi - \bxi \big\|^2 &= O_p(n_1^{- 2 \kappa_1}), \\
	\bE_{(\Y, \Z)} \big\| \hat \boldeta - \boldeta \big\|^2 &= O_p(n_1^{- 2 \kappa_2}).
\end{align*}

We provide a detailed discussion of Assumption~\ref{con:converge_velocity_field} in
the Supplementary Materials \ref{sec:proof:con:converge_velocity_field}.
Intuitively, if the $L_2$-risk of the estimated transport maps is asymptotically ignorable, the inference based on independence measure is valid.

At the population level, distance covariance admits a decomposition analogous to Pearson's covariance:
\begin{align*}
	\dcov^2(\U, \V) = \cov(\|\U_1-\U_2\|, \|\V_1-\V_2\|) - 2 \cov (\|\U_1-\U_2\|, \|\V_1-\V_3\|).
\end{align*}
This structure inherits the \emph{double robustness} property: the product of estimation rates $n_1^{-\kappa_1}$ and $n_1^{-\kappa_2}$ yields faster combined convergence \citep{chernozhukovDouble2018, liuDouble2021, shiDouble2021}.

However, unlike classical Pearson-type statistics, under $H_0$ where $\U \indep \V$, the empirical distance correlation becomes a \emph{degenerate} V-statistic.
This degeneracy yeilds an $n^{-1}$ convergence rate instead of the standard $n^{-1/2}$.
To address this, we employ a data splitting procedure \citep{zhangClassification2023}, allocating different orders to $n_1$ and $n_2$.
Thus, based on Assumption~\ref{con:converge_velocity_field}, we have the following Theorem~\ref{thm:asymptotics} regarding the asymptotic behavior of our test statistic $T_n$ from \eqref{eq:T_n}.

\begin{Theorem}
\label{thm:asymptotics}
	Suppose Assumption~\ref{con:converge_velocity_field} holds and the condition vector $\Z$ has a finite second moment. Let $n_2 = o\big(n_1^{\kappa_1+\kappa_2}\big)$. As $n_1, n_2 \to \infty$, the following convergence holds:
	\begin{enumerate}
		\item Under $H_0$, $n_2 T_{n_2} \stackrel{d}{\to} \sum_{j=1}^\infty \lambda_j Z_j^2$.
		\item Under $H_1$, $n_2 T_{n_2} \stackrel{p}{\to} \infty$.
	\end{enumerate}
	Here, $Z_j$ are independent standard normal random variables, and $\{\lambda_j\}$ are nonnegative constants determined by the distribution of $\U$ and $\V$.
\end{Theorem}

If $\bxi$ and $\boldeta$ are fully observed without estimation, the convergence of the empirical distance correlation is established in \cite{szekelyMeasuring2007}.
In our setting, however, the velocity fields are estimated via deep neural networks, and their convergence rates $\kappa_1, \kappa_2 \in (0, 1/2)$ depend on the smoothness and dimensionality of the velocity fields under nonparametric assumptions.
As the smoothness index grows to infinity, we have $\kappa_1, \kappa_2 \to 1/2$, which simplifies the rate condition to $n_2 = o(n_1)$. 
Otherwise, if $\kappa_1 + \kappa_2 > 1 / 2$, which can be achievable \citep{chernozhukovDouble2018, shiDouble2021}, the requirement becomes $n_2 = O(\sqrt{n_1})$.

In practice, we advocate allocating $n_2 \propto \sqrt{n}$. This choice balances theoretical validity with finite-sample performance, effectively controlling type-I error as demonstrated in extensive simulations.

Since the limiting distribution under $H_0$ is a mixed $\chi^2$-distribution, which is intractable, we propose using permutation to obtain a $p$-value as given in \eqref{eq:permuted_p}. The following Theorem~\ref{thm:permutation_p} ensures the validity of this approach.

\begin{Theorem}
\label{thm:permutation_p}
Suppose that the conditions in Theorem~\ref{thm:asymptotics} are satisfied.
As $n_1, n_2, B \to \infty$:
\begin{enumerate}
	\item Under $H_0$, $p_B \stackrel{d}{\to} U \sim \dU(0,1)$.
	\item Under $H_1$, $p_B \stackrel{p}{\to} 0$.
\end{enumerate}
\end{Theorem}

The permutation test is a classical approach that has been widely used in both statistics and modern machine learning literature \citep{xuNonparametric2022, caiAsymptotic2024}. Under $H_0$, the permuted statistic shares the same limiting distribution as the original test statistic. Consequently, the limiting distribution of the $p$-value is standard uniform, ensuring the validity of the test.
Proofs of Theorems~\ref{thm:asymptotics} and \ref{thm:permutation_p} are given in Sections~\ref{sec:proof:thm:asymptotics} and~\ref{sec:proof:thm:permutation_p}, respectively.
While sample splitting ensures theoretical validity, it incurs power loss. We mitigate this in Section~\ref{sec:multiple_splitting} through multiple splitting aggregation.

\subsection{Combining Results from Multiple Splitting}
\label{sec:multiple_splitting}

In this section, we adopt a multiple splitting strategy analogous to the classical $K$-fold approach. Specifically, we split the data into $\lfloor n/n_2 \rfloor$ disjoint folds.
At each iteration, we use one fold as the test set and the remaining folds for training, and compute a $p$-value following Algorithm~\ref{alg:permutation_test}.
To aggregate $m \le \lfloor n/n_2 \rfloor$ $p$-values $p_{B,1}, \ldots, p_{B, m}$, we employ the Cauchy combination method \citep{liuCauchy2020}:
\begin{align*}
	T = \sum_{i=1}^m \frac{1}{m}\tan\{(0.5 - p_{B, i}) \pi\}.
\end{align*}

When $p_{B, i}$ follows a standard uniform distribution, the transformation $\tan\{(0.5 - p_{B, i}) \pi\}$ follows a standard Cauchy distribution. Under independence of $p_{B, i}$, their average $T$ preserves the standard Cauchy distribution.
We then transform $T$ back to yield
\begin{align}
	\label{eq:p_c}
	p_{c} = 0.5 - (\arctan T)/\pi.
\end{align}
Since the $p$-values are obtained from disjoint test sets and the estimation error from the training set is ignorable based on Theorem~\ref{thm:asymptotics}, we have the following Corollary~\ref{cor:multiple_split}.

\begin{Corollary}
	\label{cor:multiple_split}
	Let $m \le \lfloor n/n_2 \rfloor$ and the test folds be disjoint.
	Then, the combined $p$-value $p_c$ given in \eqref{eq:p_c} satisfies:
	\begin{enumerate}
		\item Under $H_0$, $p_c \stackrel{d}{\to} U \sim \dU(0,1)$.
		\item Under $H_1$, $p_c \stackrel{p}{\to} 0$.
	\end{enumerate}
\end{Corollary}

\section{Simulations}
\label{sec:sim}

We  validate the proposed method through extensive simulations.
In Section~\ref{sec:convergence_simulations}, we focus on demonstrating the convergence of the $p$-values under $H_0$, as given in Theorem~\ref{thm:permutation_p} and Corollary~\ref{cor:multiple_split}.
We will present the empirical type-I error and statistical power performance, as well as comparisons with other tests, in Section~\ref{sec:comparison_simulations}.

\subsection{Convergence of the Test Statistic}
\label{sec:convergence_simulations}

\begin{Model}[Convergence under $H_0$]
	\label{mod:convergence}
	Dimensions $(d_\x, d_\y, d_\z)$ and sample size $n$ vary across experiments.
	Data generation proceeds as follows:
	\begin{enumerate}
		\item Generate parameter matrices $\B_1 \in \bR^{d_\z \times d_\x}$ and $\B_2 \in \bR^{d_\z \times d_\y}$, where each element is independently drawn from $\dN(0, 1)$.
		\item Generate error terms $\beps_{\x} \sim \dN(\bzero, \I_{d_\x})$ and $\beps_{\y} \sim \dN(\bzero, \I_{d_\y})$.
		\item Generate $\Z \sim \dN(\bzero, \I_{d_\z})$, then compute $\X=\Z \B_1 + \beps_\x$ and $\Y = \Z \B_2 + \beps_\y$.
	\end{enumerate}
\end{Model}

\begin{figure}[ht]
	\centering
	\subfloat[\label{fig:sub_model0_d1_n500_m1}]{
		\includegraphics[width=0.245\linewidth]{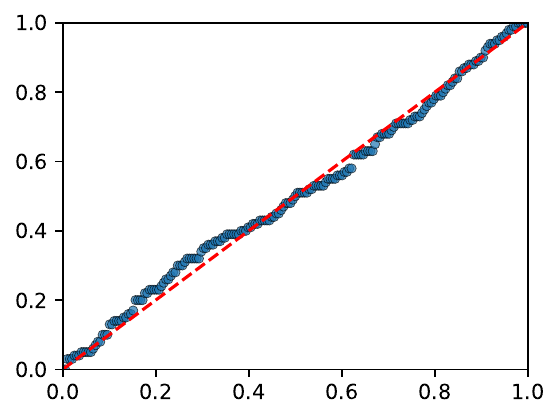}}
	\subfloat[\label{fig:sub_model0_d1_n1000_m1}]{
		\includegraphics[width=0.245\linewidth]{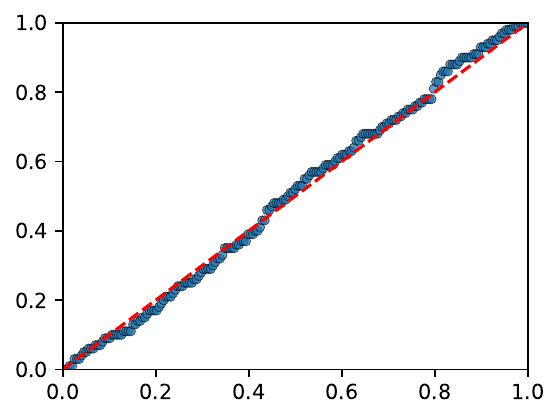}}
	\subfloat[\label{fig:sub_model0_d5_n500_m1}]{
		\includegraphics[width=0.245\linewidth]{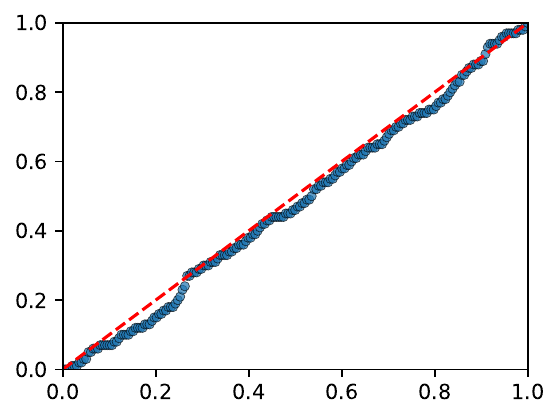}}
	\subfloat[\label{fig:sub_model0_d5_n1000_m1}]{
		\includegraphics[width=0.245\linewidth]{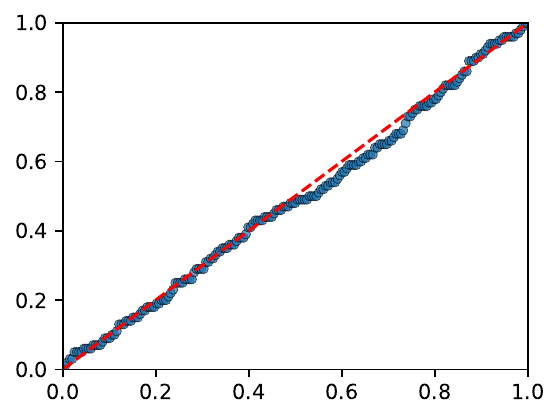}}
	\caption{
		Finite-sample performance of Theorem~\ref{thm:permutation_p} under $H_0$.
		Q-Q plots compare empirical $p$-values against the standard uniform distribution for Model~\ref{mod:convergence}.  (a) $(d_\x, d_\y, d_\z)=(1, 1, 1)$ and $n=500$; (b) $(d_\x, d_\y, d_\z)=(1, 1, 1)$ and $n=1000$;
(c) $(d_\x, d_\y, d_\z)=(5, 5, 5)$ and $n=500$;  and (d) $(d_\x, d_\y, d_\z)=(5, 5, 5)$ and $n=1000$.
Results are based on 200 replications.
	}
	\label{fig:model0_m1}
\end{figure}

To perform the test, we split the data with $n_2 = \lfloor 4 \sqrt{n} \rfloor$, following the recommendations that $n_2 \propto \sqrt{n}$.
We test the hypothesis $\boldeta \indep (\bxi, \Z)$ under two configurations:
$(d_\x, d_\y, d_\z)=(1, 1, 1)$ and $(5, 5, 5)$, with sample sizes $n=500$ and $n=1000$.

Figure~\ref{fig:model0_m1} displays Q-Q plots comparing empirical $p$-values to the standard uniform distribution. Each subplot corresponds to different dimension and sample size combinations. The close alignment between empirical $p$-values and the theoretical uniform distribution supports the validity of our asymptotic results.

We further validate the multiple splitting procedure from Section~\ref{sec:multiple_splitting} by repeating the split $m=5$ times.
The corresponding results, shown in the Supplementary Materials \ref{appendix:sec:multiple_split_H0} (Figure~\ref{fig:model0_m5}).
mirror the patterns observed in Figure~\ref{fig:model0_m1}. These findings collectively suggest that Theorem~\ref{thm:permutation_p} and Corollary~\ref{cor:multiple_split} hold in finite samples.

\subsection{Comparisons with Existing Methods}
\label{sec:comparison_simulations}

In this section, we compare our test with the following nonparametric conditional independence test methods:
kernel-based conditional independence test (KCI, \citealt{zhangKernelbased2011}),
conditional distance correlation test (CDC, \citealt{wangConditional2015}),
classifier conditional independence test (CCIT, \citealt{senModelpowered2017}),
fast conditional independence test (FCIT, \citealt{chalupkaFast2018}),
a test proposed by \cite{zhouTest2020} (abbreviated as ZLZ),
and a test proposed by \cite{caiDistribution2022} (abbreviated as CLZ).
The KCI test is implemented by the \texttt{R} package \texttt{CondIndTests::KCI},
the CDC and FCIT are implemented by the \texttt{Python} package \texttt{hyppo} \citep{pandaHyppo2024},
the CCIT is implemented by its \texttt{Python} package \texttt{CCIT},
the ZLZ test is conducted based on its original \texttt{Matlab} code,
and the CLZ test is implemented by its original \texttt{R} code.
Please refer to the Supplementary Materials \ref{sec:implement}
for the details of implementations of these methods.

We consider four different examples in Models~\ref{mod:univariate}, \ref{mod:model_low_low}, \ref{mod:model_low_high}, and \ref{mod:model_high_high}. These models range from linear to nonlinear examples and include univariate, low-dimensional, and moderately high-dimensional scenarios.

As a prelude, we summarize the comparisons of these methods based on our simulation models in Table~\ref{tab:sim_comparisons}. Additionally, our method effectively controls the type-I error under $H_0$ and achieves high statistical power performance across all simulation models.

\medskip
\begin{table}[H]
\medskip
\centering
\caption{Comparisons of nonparametric conditional independence test methods across our simulation settings. A checkmark ($\newcheckmark$) indicates that the method is workable under the given setting---effectively controlling the type-I error, demonstrating statistical power against alternatives, and being computationally feasible.
A triangle ($\Delta$) indicates that the method is partially workable.
Cells are left blank if a method fails to show detectable statistical power under our model settings or is computationally infeasible in practice.
}
\label{tab:sim_comparisons}
\begin{tabular}{c c c c c c c c c c}
\toprule
\multicolumn{2}{c}{Dimensions} & \multirow{2}{*}{Examples} & \multicolumn{1}{c}{\multirow{2}{*}{FlowCIT}} & \multicolumn{1}{c}{\multirow{2}{*}{KCI}} & \multicolumn{1}{c}{\multirow{2}{*}{CDC}} & \multicolumn{1}{c}{\multirow{2}{*}{FCIT}} & \multicolumn{1}{c}{\multirow{2}{*}{CCIT}} & \multicolumn{1}{c}{\multirow{2}{*}{ZLZ}} & \multicolumn{1}{c}{\multirow{2}{*}{CLZ}} \\
$\Z$ & $\X$ and $\Y$ &  & \multicolumn{1}{c}{} & \multicolumn{1}{c}{} & \multicolumn{1}{c}{} & \multicolumn{1}{c}{} & \multicolumn{1}{c}{} & \multicolumn{1}{c}{} \\
\hline
\multirow{2}{*}{Low} & Univariate & Model~\ref{mod:univariate} & $\newcheckmark$ & $\newcheckmark$ & $\newcheckmark$ & $\newcheckmark$ & $\newcheckmark$ & $\newcheckmark$ & $\newcheckmark$ \\
 & Low & Model~\ref{mod:model_low_low} & $\newcheckmark$ & $\newcheckmark$ & $\newcheckmark$ & {$\Delta$} &  &  \\
\hline
\multirow{2}{*}{High} & Low & Model~\ref{mod:model_low_high} & $\newcheckmark$ &  &  & {$\Delta$} & {$\Delta$} &  \\
 & High & Model~\ref{mod:model_high_high} & $\newcheckmark$ &  &  & {$\Delta$} & {$\Delta$} &  \\
\bottomrule
\end{tabular}
\end{table}

In these models, we use the parameter $\psi$ to control the deviation from $H_0$: if $\psi=0$, the model is under $H_0$; otherwise, a larger $|\psi|$ indicates a greater deviation.
We use plots to visualize the empirical type-I errors and statistical powers across various settings.

Since the data points are densely clustered under $H_0$, detailed empirical type-I error rates are provided in
Supplementary Material \ref{appendix:sec:typeIerror}.
In this section, we implement our method using distance correlation to test the independence of $\boldeta$ and $(\bxi, \Z)$. We also observe that testing the independence of $\bxi$ and $(\boldeta, \Z)$ yields similar performance in
Appendix \ref{appendix:sec:compare_two_formula}.
A comparison between distance correlation and improved projection correlation is presented in Appendix \ref{appendix:sec:compare_DC_IPC}, indicating that their performances are comparable. All results are based on 200 replications and the data is split with $n_2 = \lfloor 4 \sqrt{n} \rfloor$.

\subsubsection{Low-dimensional $\Z$}
\label{sec:sim:low-dimensional-Z}

We start with examples involving low-dimensional $\Z$.
A univariate $X$ and $Y$ example (Model~\ref{mod:univariate}) is deferred to
Supplementary Materials \ref{appendix:sec:univariate_x_y}. In the following Model~\ref{mod:model_low_low}, we analyze multivariate pairs $\X$ and $\Y$, with all three vectors specified as three-dimensional. The multiple splitting is repeated $m=5$ times in this section.

\begin{Model}[Multivariate $\X$, $\Y$, and $\Z$]
\label{mod:model_low_low}
Let $(d_\x, d_\y, d_\x)=(3,3,3)$ and the sample size $n=500$.
Each element of $\B_1$, $\B_2$, $\B_3$, and $\Z$ are generated from $\dN(0, 1)$.
Vectors $\X$ and $\Y$ are generated under Settings~\ref{set:low_low-1}--\ref{set:low_low-4}, and we vary $\psi$ over the set $\{0, 0.05, 0.1, 0.15, 0.2\}$.
For the error terms, we generate $\beps_\y \sim \dN(\bzero, \I_{d_\y})$, and $\beps_\x \sim \dN(\bzero, \I_{d_\x})$.
\begin{Setting}
\label{set:low_low-1}
$\X = \Z \B_1 + \beps_\x, \quad \Y = \Z \B_2 + \psi \X \B_3 + \beps_\y$.
\end{Setting}
\begin{Setting}
\label{set:low_low-2}
$\X = \Z \B_1 + \beps_\x, \quad \Y = \sin(\Z \B_2) + \psi \X \B_3 + \beps_\y$.
\end{Setting}
\begin{Setting}
\label{set:low_low-3}
$\X = (\Z \B_1)^2 + \beps_\x, \quad \Y = \Z \B_2 + \psi \X \B_3 + \beps_\y$.
\end{Setting}
\begin{Setting}
\label{set:low_low-4}
$\X = \Z \B_1 + \beps_\x, \quad \Y = \Z \B_2 + |\psi \X \B_3| + \beps_\y$.
\end{Setting}
\end{Model}

\begin{figure}[ht]
\centering
\subfloat[Setting~\ref{set:low_low-1}\label{fig:sub_model_low_low_n500_s1}]{
\includegraphics[width=0.48\linewidth]{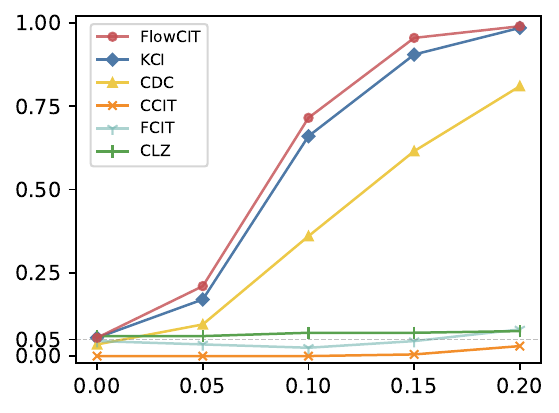}}
\hfil
\subfloat[Setting~\ref{set:low_low-2}\label{fig:sub_model_low_low_n500_s2}]{
\includegraphics[width=0.48\linewidth]{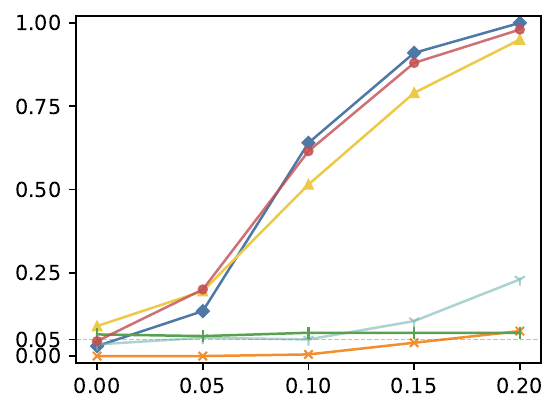}}
\vfil
\subfloat[Setting~\ref{set:low_low-3}\label{fig:sub_model_low_low_n500_s3}]{
\includegraphics[width=0.48\linewidth]{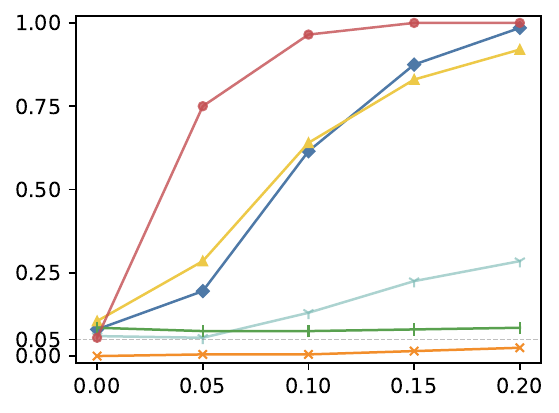}}
\hfil
\subfloat[Setting~\ref{set:low_low-4}\label{fig:sub_model_low_low_n500_s4}]{
\includegraphics[width=0.48\linewidth]{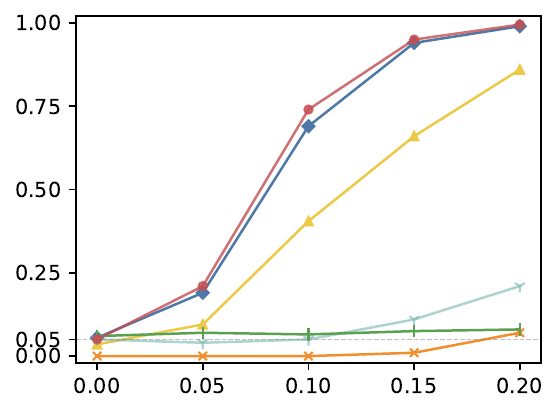}}
\caption{
Empirical type-I errors and statistical powers of Model~\ref{mod:model_low_low} with sample size $n=500$.
The dashed gray line below refers to significance level of $0.05$.
The $x$-axis refers to the value of $\psi$.
}
\label{fig:model_low_low_n500}
\end{figure}

The results for Model~\ref{mod:model_low_low} are shown in Figure~\ref{fig:model_low_low_n500}, with the $x$-axis corresponding to $\psi$.
Since ZLZ is designed for univariate $X$ and $Y$, we focus on comparing the remaining six methods.
Under $H_0$, all six methods effectively control the type-I error.
However, under $H_1$, CCIT and CLZ demonstrate reduced statistical power.
FCIT shows limited statistical power in the linear Setting~\ref{set:low_low-1} but achieve modest statistical power in the nonlinear Settings~\ref{set:low_low-2}, \ref{set:low_low-3}, and \ref{set:low_low-4}, though still under-performing FlowCIT, KCI, and CDC.
Our proposed FlowCIT consistently attains the highest or comparable statistical power to KCI and CDC across all settings.

\subsubsection{Moderately high-dimensional $\Z$}
\label{sec:sim:high-dimensional-Z}

Since CLZ and KCI are kernel-based methods that struggle in high-dimensional settings due to computational burdens, we limit our comparison in Models~\ref{mod:model_low_high}--\ref{mod:model_high_high} to CDC, FCIT, and CCIT.
The multiple splitting is repeated $m=2$ times in this section.

In high-dimensional $\Z$ examples (Models~\ref{mod:model_low_high}--\ref{mod:model_high_high}), we first consider a low-dimensional pair $(\X, \Y)$ in Model~\ref{mod:model_low_high}, and then increase the dimensions of $(\X, \Y)$ in Model~\ref{mod:model_high_high}.
In these models, our test demonstrates higher statistical power compared to other methods. FCIT and CCIT are effective in certain settings but ineffective in others.

\begin{Model}[Low-dimensional $\X$ and $\Y$, and moderately high-dimensional $\Z$]
\label{mod:model_low_high}
Let $(d_\x, d_\y, d_\z) = (5, 5, 50)$ and the sample size $n=1000$.
Each element of $\beps_\x$, $\beps_\y$, and $\Z$ is generated from $\dN(0, 1)$.
Matrices $\B_1$, $\B_2$, $\B_3$, and vectors $\X$ and $\Y$ are generated under Settings~\ref{set:low_high-1}--\ref{set:low_high-4}, and the parameter $\psi$ is varied over the set $\{0, 0.1, 0.2, 0.3, 0.4\}$.

\begin{Setting}[Row-sparse $\B_1$ and $\B_2$, and dense $\B_3$]
\label{set:low_high-1}
Each element of $\B_3$, as well as the first $3 \times d_\x$ sub-matrix of $\B_1$, and the first $3 \times d_\y$ sub-matrix of $\B_2$, is generated from $\dN(0, 1)$.
All other elements of $\B_1$ and $\B_2$ are set to zero.
We then generate:
$\X = \Z \B_1 + \beps_\x, \quad \Y = \Z \B_2 + \psi \X \B_3 + \beps_\y.$
\end{Setting}

\begin{Setting}[Sparse $\B_1$, $\B_2$, and $\B_3$]
\label{set:low_high-2}
Each element of the first $3 \times 1$ sub-matrices of $\B_1$, $\B_2$, and $\B_3$ is generated from $\dN(0, 1)$.
All other elements of $\B_1$, $\B_2$, and $\B_2$ are set to zero.
We then generate:
$\X = \Z \B_1 + \beps_\x, \quad \Y = (\Z \B_2)^2 + 4\psi \X \B_3 + \beps_\y.$
\end{Setting}

\begin{Setting}[Row-sparse $\B_1$ and $\B_2$, and dense $\B_3$]
\label{set:low_high-3}
Each element of $\B_3$, as well as the first $2 \times d_\x$ sub-matrix of $\B_1$, and the first $2 \times d_\y$ sub-matrix of $\B_2$, is generated from $\dN(0, 1)$.
All other elements of $\B_1$, $\B_2$, and $\B_3$ are set to zero.
We then generate:
$\X = \Z \B_1 + \beps_\x, \quad \Y = \Z \B_2 + \psi \X \B_3 + \beps_\y.$
\end{Setting}

\begin{Setting}[Dense $\B_1$, $\B_2$, and $\B_3$]
\label{set:low_high-4}
Each element of $\B_1$, $\B_2$, and $\B_3$ is generated from $\dN(0,1)$. We then generate:
$\X = \sin(\Z \B_1) + \beps_\x, \quad \Y = \Z \B_2 + |5\psi \X \B_3| + \beps_\y.$
\end{Setting}
\end{Model}

\begin{figure}[!htb]
\centering
\subfloat[Setting~\ref{set:low_high-1}\label{fig:sub_model_low_high_n1000_s1}]{
\includegraphics[width=0.48\linewidth]{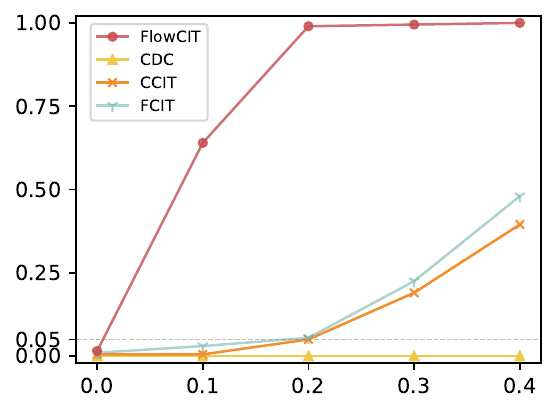}}
\hfil
\subfloat[Setting~\ref{set:low_high-2}\label{fig:sub_model_low_high_n1000_s2}]{
\includegraphics[width=0.48\linewidth]{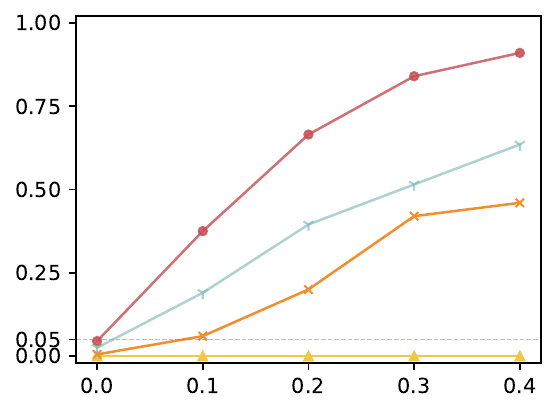}}
\vfil
\subfloat[Setting~\ref{set:low_high-3}\label{fig:sub_model_low_high_n1000_s3}]{
\includegraphics[width=0.48\linewidth]{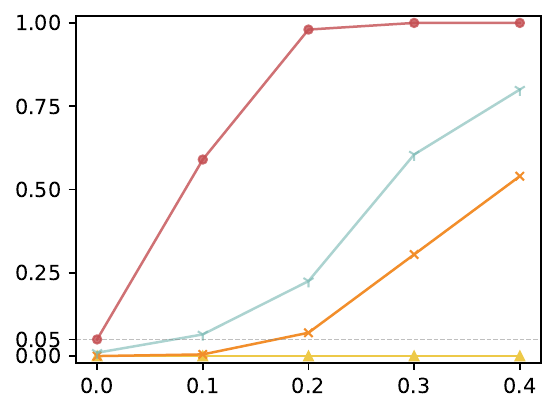}}
\hfil
\subfloat[Setting~\ref{set:low_high-4}\label{fig:sub_model_low_high_n1000_s4}]{
\includegraphics[width=0.48\linewidth]{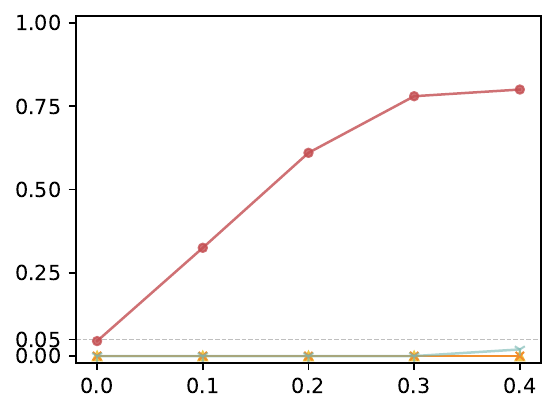}}
\caption{
Empirical type-I errors and statistical powers of Model~\ref{mod:model_low_high} with sample size $n=1000$.
The dashed gray line below refers to significance level of $0.05$.
The $x$-axis refers to the value of $\psi$.
}
\label{fig:model_low_high_n1000}
\end{figure}

The results for Model~\ref{mod:model_low_high} are shown in Figure~\ref{fig:model_low_high_n1000}. In sparse settings (Settings~\ref{set:low_high-1}--\ref{set:low_high-3}), our method demonstrates robust performance. FCIT and CCIT also achieve statistical power under the alternative hypothesis, though they underperform compared to our method. In the dense setting (Setting~\ref{set:low_high-4}), our method retains its effectiveness, while FCIT and CCIT show no statistical power under the alternative hypothesis.

In the next Model~\ref{mod:model_high_high}, we increase the dimension of $\X$ and $\Y$, considering settings with $(d_\x, d_\y, d_\z)=(50, 50, 50)$.

\begin{Model}[Moderately high-dimensional $\X$, $\Y$, and $\Z$]
\label{mod:model_high_high}
Let $(d_\x, d_\y, d_\z) = (50, 50, 50)$ and the sample size $n=1000$.
Each element of $\beps_\x$, $\beps_\y$, and $\Z$ is generated from $\dN(0, 1)$.
Matrices $\B_1$, $\B_2$, $\B_3$, and vectors $\X$ and $\Y$ are generated under Settings~\ref{set:high_high-1}--\ref{set:high_high-4}, and the parameter $\psi$ is varied over the set $\{0, 0.2, 0.4, 0.6, 0.8\}$.

\begin{Setting}[Row-sparse $\B_1$ and $\B_2$, and dense $\B_3$]
\label{set:high_high-1}
Each element of $\B_3$ and the first two rows of $\B_1$ and $\B_2$ is generated from $\dN(0, 1)$.
All other elements of $\B_1$ and $\B_2$ are set to zero.
The sparsity percentages of $\B_1$ and $\B_2$ are $4\%$.
We then generate:
$$\X = \Z \B_1 + \beps_\x, \quad \Y = \Z \B_2 + \psi \X \B_3 + \beps_\y.$$
\end{Setting}

\begin{Setting}[Row-sparse $\B_1$ and $\B_2$, and dense $\B_3$]
	\label{set:high_high-2}
	Each element of $\B_3$ and the first row of $\B_1$ and $\B_2$ is generated from $\dN(0, 1)$.
	The sparsity percentages of $\B_1$ and $\B_2$ are $2\%$.
	The other setttings are the same as Setting~\ref{set:high_high-1}.
\end{Setting}

\begin{Setting}[Sparse $\B_1$, $\B_2$, and $\B_3$]
\label{set:high_high-3}
Each element of the first $3\times 3$ sub-matrices of $\B_1$, $\B_2$, and $\B_3$ is generated from $\dN(0, 1)$.
All other elements of $\B_1$, $\B_2$, and $\B_2$ are set to zero.
The sparsity percentages of $\B_1$, $\B_2$, and $\B_3$ are $0.18\%$, $0.18\%$, and $0.36\%$.
We then generate:
$$\X = \Z \B_1 + \beps_\x, \quad \Y = \Z \B_2 + |\psi \X \B_3|  + \beps_\y.$$
\end{Setting}

\begin{Setting}[Sparse $\B_1$, $\B_2$, and $\B_3$]
\label{set:high_high-4}
Each element of matrices $\B_1$, $\B_2$, and $\B_3$ is randomly taken 0 with probability $90\%$ and generated from $\dN(0, 1)$ with probability $10\%$.
The expectation of the sparsity percentages of these matrices are $10\%$. We then generate:
$$\X = \cos(\Z \B_1) + \beps_\x, \quad \Y = \sin(\Z \B_2) + \psi \X \B_3 + \beps_\y.$$
\end{Setting}
\end{Model}

\begin{figure}[!htb]
\centering
\subfloat[Setting~\ref{set:high_high-1}\label{fig:sub_model_high_high_n1000_s1}]{
\includegraphics[width=0.48\linewidth]{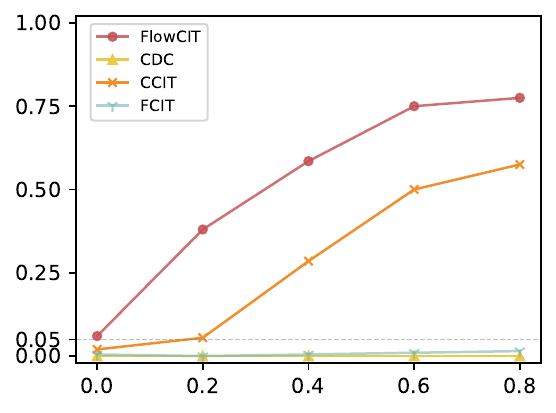}
}
\hfil
\subfloat[Setting~\ref{set:high_high-2}\label{fig:sub_model_high_high_n1000_s2}]{
\includegraphics[width=0.48\linewidth]{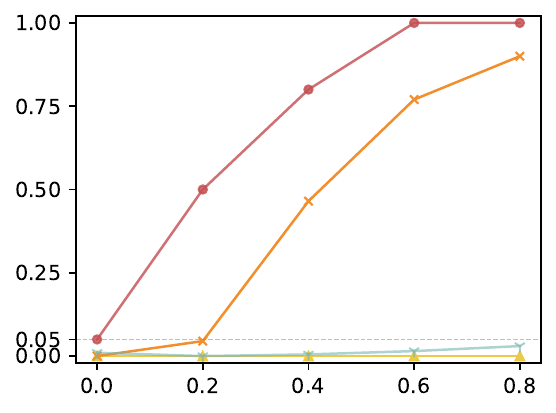}
}
\vfil
\subfloat[Setting~\ref{set:high_high-3}\label{fig:sub_model_high_high_n1000_s3}]{
\includegraphics[width=0.48\linewidth]{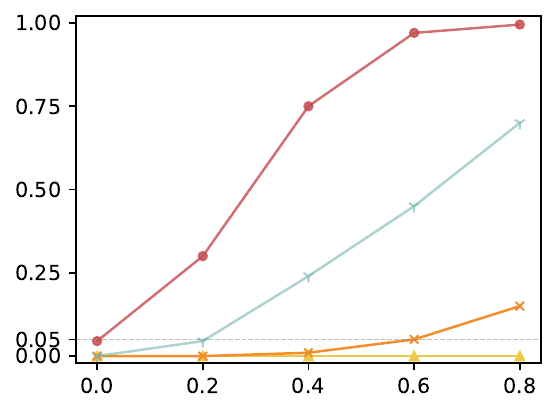}
}
\hfil
\subfloat[Setting~\ref{set:high_high-4}\label{fig:sub_model_high_high_n1000_s4}]{
\includegraphics[width=0.48\linewidth]{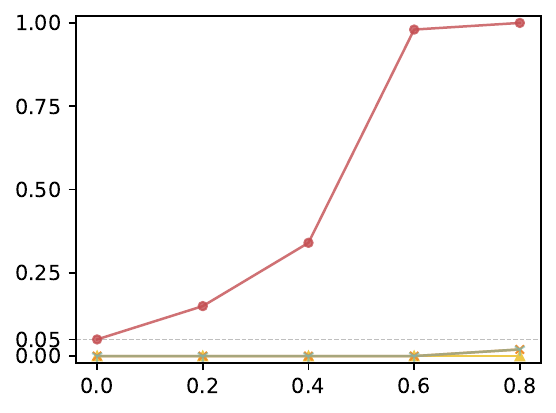}
}
\caption{
Empirical type-I errors and statistical powers of Model~\ref{mod:model_high_high} with sample size $n=1000$.
The dashed gray line below refers to significance level of $0.05$.
The $x$-axis refers to the value of $\psi$.
}
\label{fig:model_high_high_n1000}
\end{figure}

The results for Model~\ref{mod:model_high_high} are shown in Figure~\ref{fig:model_high_high_n1000}.
Our method effectively controls the type-I error and demonstrates greater statistical power compared to other methods.
FCIT is partially effective in the scenarios with sparse $\B_1$, $\B_2$, and $\B_3$ when the sparsity percentage is below $1\%$ (Setting~\ref{set:high_high-3}).
CCIT is partially workable in scenarios with row-sparse $\B_1$ and $\B_2$ and dense $\B_3$ (Setting~\ref{set:high_high-1} and Setting~\ref{set:high_high-2}).
When the sparsity percentage increases to around $10\%$ (Setting~\ref{set:high_high-4}), our proposed FlowCIT is more effective.

\section{Application to Dimension Reduction}
\label{sec:real_data}

Conditional independence testing holds broad applicability across statistical methodologies.
We demonstrate its practical utility through real-data applications in dimension reduction.

Dimension reduction is a cornerstone of modern statistics and machine learning.
Given paired data $(\Y, \X)$, the goal is to identify a low-dimensional representation $\calR(\X)$---linear or nonlinear---that satisfies the conditional independence criterion:
$\Y \indep \X \mid \calR(\X).$
The efficacy of this reduction hinges critically on both the dimensionality and function form of $\calR(\X)$.
To rigorously evaluate whether a candidate $\calR(\X)$ achieves sufficiency, we can leverage our conditional independence test framework.

We evaluate dimension reduction methods using the wine quality dataset from \citet{cortezModeling2009}, available at \url{https://archive.ics.uci.edu/dataset/186/wine+quality}.
The dataset includes $n = 4898$ samples of white wine, with predictors (fixed acidity, volatile acidity, citric acid, residual sugar, chlorides, free sulfur dioxide, total sulfur dioxide, density, pH, sulfates, alcohol) and a response variable (wine quality rating).

Our goal is to test whether a dimension reduction method, represented as $\calR(\X),$  satisfies the conditional independence $\Y \indep \X \mid \calR(\X)$. We evaluate four dimension reduction methods:
\begin{enumerate}
    \item Principal Component Analysis (PCA, \citealt{pearsonLines1901}), an unsupervised linear method;
    \item Sliced Inverse Regression (SIR, \citealt{liSliced1991}), a supervised linear method;
    \item Uniform Manifold Approximation and Projection (UMAP, \citealt{mcinnesUMAP2018}), an unsupervised nonlinear method;
    \item Deep Dimension Reduction (DDR, \citealt{huangDeep2024}), a supervised nonlinear method.
\end{enumerate}

For each method, we test reduced dimensions $d = 1$ and $d = 4$. The data are split into training (90\%) and testing (10\%) subsets: $\calR(\X)$ is learned on the training set, and conditional independence is tested on the held-out set. As a baseline, we set $\calR(\X) = \X$, which trivially satisfies $\Y \indep \X \mid \X$.

\begin{table}[H]
\centering
\medskip
\caption{$p$-values of conditional independence test of $\Y \indep \X \mid \calR(\X)$ after learning $\calR(\X)$ via specific methods. Bold: $p \le 0.05$; Underlined: $0.05 < p \le 0.1$.}
\label{tab:real_data_wine}
\begin{tabular}{r c c c c c c c c}
\toprule
Methods & Supervised & Nonlinear & FlowCIT & KCI & CDC & CCIT & FCIT & CLZ \\
\midrule
PCA-1 & $\xmark$ & $\xmark$ & \textbf{0.00} & \textbf{0.04} & \textbf{0.01} & \underline{0.09} & \textbf{0.00} & \textbf{0.00} \\
PCA-4 & $\xmark$ & $\xmark$ & \textbf{0.00} & \textbf{0.00} & \textbf{0.01} & 0.68 & 0.27 & \textbf{0.00} \\
SIR-1 & $\newcheckmark$ & $\xmark$ & \textbf{0.04} & \underline{0.09} & \textbf{0.02} & 0.19 & 0.94 & \textbf{0.00} \\
SIR-4 & $\newcheckmark$ & $\xmark$ & \textbf{0.02} & \textbf{0.00} & \textbf{0.05} & 0.27 & 0.79 & \underline{0.07} \\
UMAP-1 & $\xmark$ & $\newcheckmark$ & \textbf{0.00} & \underline{0.08} & \textbf{0.01} & 0.29 & 0.22 & \textbf{0.00} \\
UMAP-4 & $\xmark$ & $\newcheckmark$ & \textbf{0.00} & \textbf{0.00} & \textbf{0.01} & 0.37 & \textbf{0.01} & \textbf{0.00} \\
DDR-1 & $\newcheckmark$ & $\newcheckmark$ & \textbf{0.01} & 0.12 & 0.13 & 0.32 & 0.59 & 0.99 \\
DDR-4 & $\newcheckmark$ & $\newcheckmark$ & 0.30 & 0.14 & 0.38 & 0.53 & 0.82 & 0.81 \\
\midrule
$\calR(\X) = \X$ &  &  & 0.71 & 0.52 & 0.28 & 0.68 & 0.18 & 1.00 \\
\bottomrule
\end{tabular}
\end{table}

Results are summarized in Table~\ref{tab:real_data_wine}. Under the baseline ($\calR(\X) = \X$), all tests fail to reject $H_0$, confirming the validity of the testing framework.
For PCA, SIR, and UMAP, four tests (FlowCIT, KCI, CDC, and CLZ) reject $H_0$ at $\alpha = 0.1$, indicating $\Y \not\indep \X \mid \calR(\X)$ and suggesting these dimension reduction methods inadequately preserve sufficient information.
In contrast, the supervised nonlinear DDR method with $d = 4$ satisfies $\Y \indep \X \mid \calR(\X)$ across all tests.
In addition, DDR with $d = 1$ is rejected by FlowCIT, which suggests that reducing predictors into one dimension is not sufficient.

\section{Conclusion}
\label{sec:conclusion}

In this work, we propose a novel framework for conditional independence testing that utilizes transport maps. By constructing two transport maps using conditional continuous normalizing flows, we transform the conditional independence problem into an unconditional independence test on the transformed variables. This transformation allows for the application of standard independence measures while preserving statistical validity. Numerical experiments demonstrate the framework's superior performance across a variety of scenarios, including univariate, low-dimensional, and high-dimensional settings, and demonstrate its robustness to the choice of dependence measures.

In future work, we plan to explore the use of alternative transport maps for converting conditional independence problems into unconditional ones. We will also consider applying the proposed method to other formulations of conditional independence to further enhance the framework's applicability and effectiveness.

\newpage

\appendix

\setcounter{figure}{0}
\setcounter{table}{0}
\setcounter{equation}{0}
\setcounter{Theorem}{0}
\setcounter{Assumption}{0}
\setcounter{Remark}{0}
\setcounter{Model}{0}
\counterwithout{equation}{section}
\renewcommand{\thetable}{S.\arabic{table}}
\renewcommand{\thefigure}{S.\arabic{figure}}
\renewcommand{\theequation}{S.\arabic{equation}}
\renewcommand{\theTheorem}{S.\arabic{Theorem}}
\renewcommand{\theLemma}{S.\arabic{Lemma}}
\renewcommand{\theHTheorem}{S.\arabic{Theorem}}
\renewcommand{\theAssumption}{S.\arabic{Assumption}}
\renewcommand{\theHAssumption}{S.\arabic{Assumption}}
\renewcommand{\theRemark}{S.\arabic{Remark}}
\renewcommand{\theHRemark}{S.\arabic{Remark}}
\renewcommand{\theModel}{S.\arabic{Model}}
\renewcommand{\theHModel}{S.\arabic{Model}}

\begin{center}
\textbf{\LARGE APPENDIX}
\end{center}

\medskip

In the Appendix, we include additional discussions, numerical studies, and proofs for the main paper.

\section{Improved Projection Correlation}
\label{sec:ipc}

We first provide a brief introduction for the improved projection correlation (IPC, \citealt{zhangProjective2024}) in this section.
For independence copies $(\bxi_1, \boldeta_1)$, $(\bxi_2, \boldeta_2)$, and $(\bxi_3, \boldeta_3)$, recall the distance covariance defined in \eqref{eq:dc_pop_simplified}:
\begin{align*}
	\nonumber
	\dcov^2(\bxi, \boldeta) =& \bE\left( \|\bxi_1 - \bxi_2\| \| \boldeta_1 - \boldeta_2 \| \right) + \bE\|\bxi_1 - \bxi_2\| \bE \|\boldeta_1-\boldeta_2\| \\
	& - 2 \bE \left(\|\bxi_1 - \bxi_2\| \|\boldeta_1-\boldeta_3\|\right).
\end{align*}

Improved projection covariance is defined in a similar manner:
\begin{align*}
	\nonumber
	\ipcov^2(\bxi, \boldeta) =& \bE\left\{ A(\bxi_1, \bxi_2) A(\boldeta_1, \boldeta_2) \right\} + \bE\{ A(\bxi_1, \bxi_2)\} \bE \{A(\boldeta_1, \boldeta_2)\} \\
	& - 2 \bE \left\{A(\bxi_1, \bxi_2) A(\boldeta_1, \boldeta_3)\right\},
\end{align*}
where $A(\U_1, \U_2) = \arccos \{(\sigma_u^2 + \U_1^\top \U_2) (\sigma_u^2 + \U_1^\top \U_1)^{-1/2}(\sigma_u^2 + \U_2^\top \U_2)\}$ is the arc-cosine function, and $\sigma_z^2$ is suggested to choose as the median of $\U^\top \U$.
The improved projection correlation is defined as
\begin{align*}
	\ipcorr^2(\bxi, \boldeta) = \frac{\ipcov^2(\bxi, \boldeta)}{\ipcov(\bxi, \bxi) \ipcov(\boldeta, \boldeta)}.
\end{align*}

In contrast to DC, IPC imposes no moment conditions on the random vectors, rendering it more robust.
Furthermore, \cite{zhangProjective2024} demonstrates that under mild regularity conditions, the scaled statistic $n \ipcorr^2(\bxi, \boldeta)$ converges in distribution to a normal limit, thereby yielding a tractable asymptotic distribution for hypothesis testing.

\section{Additional Details of Numerical Studies}
\label{sec:additional_details_sim}

\subsection{Additional Results for the Convergence of the Test Statistic}
\label{appendix:sec:multiple_split_H0}

\begin{figure}[ht]
	\centering
	\subfloat[$(d_\x, d_\y, d_\z)=(1, 1, 1)$ and $n=500$\label{fig:sub_model0_d1_n500_m5}]{
		\includegraphics[width=0.48\linewidth]{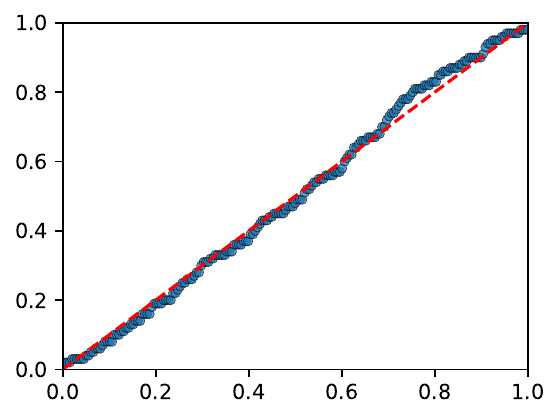}}
	\hfil
	\subfloat[$(d_\x, d_\y, d_\z)=(1, 1, 1)$ and $n=1000$\label{fig:sub_model0_d1_n1000_m5}]{
		\includegraphics[width=0.48\linewidth]{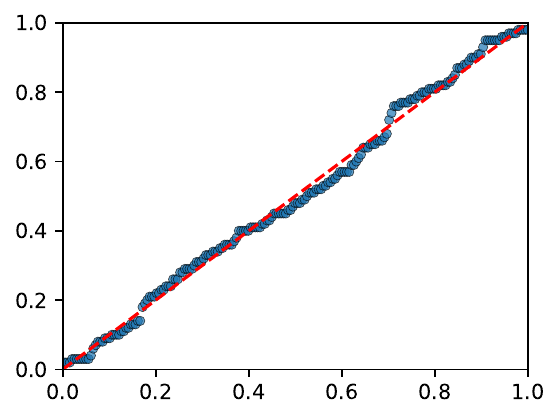}}
	\vfil
	\subfloat[$(d_\x, d_\y, d_\z)=(5, 5, 5)$ and $n=500$\label{fig:sub_model0_d5_n500_m5}]{
		\includegraphics[width=0.48\linewidth]{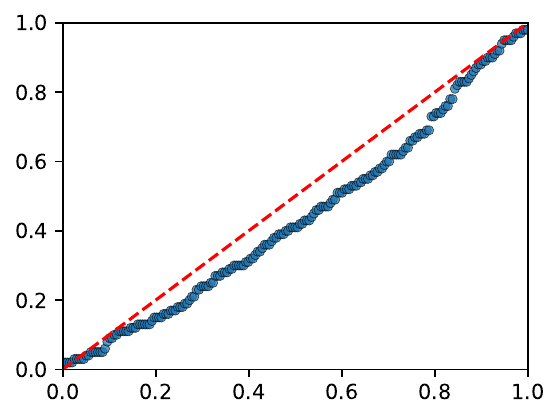}}
	\hfil
	\subfloat[$(d_\x, d_\y, d_\z)=(5, 5, 5)$ and $n=1000$\label{fig:sub_model0_d5_n1000_m5}]{
		\includegraphics[width=0.48\linewidth]{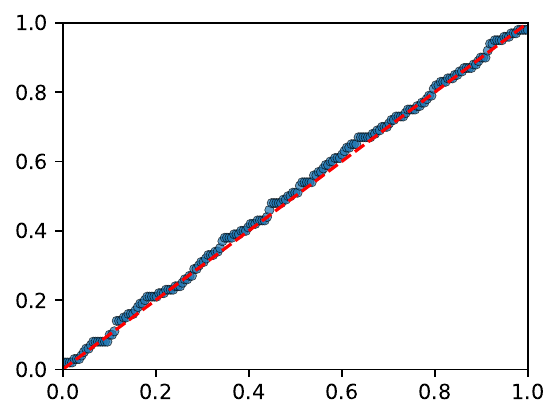}}
	\caption{
		Finite-sample performance of Corollary~\ref{cor:multiple_split} under $H_0$ with $m=5$ splits.
		Q-Q plots compare empirical $p$-values against the standard uniform distribution for Model~\ref{mod:convergence}. Results are based on 200 replications.
	}
	\label{fig:model0_m5}
\end{figure}

We provide supplementary evidence for Corollary~\ref{cor:multiple_split} under $H_0$ through Figure~\ref{fig:model0_m5}.
These results mirror the patterns observed in Figure~\ref{fig:model0_m1}, demonstrating close alignment with the standard uniform distribution.

In Figure~\ref{fig:sub_model0_d5_n500_m5}, a slight deviation from the uniform distribution is observed.
This discrepancy arises because the model operates in a relatively high-dimensional space ($d_\x + d_\y + d_\z = 15$) with a limited sample size ($n$=500), and the Cauchy combination method amplifies such deviations.
We further note that the empirical type-I error in this setting is 0.06 under the nominal significance level $\alpha=0.05$.
While this inflation exceeds the nominal level, the deviation remains acceptable in practice.
When the sample size increases to $n=1000$, the empirical $p$-value aligns closely with the standard uniform distribution.

\subsection{A Univariate $X$ and $Y$ Example}
\label{appendix:sec:univariate_x_y}

In this section, we present an additional univariate $X$ and $Y$ example comparing with several tests.
In Model~\ref{mod:univariate}, we consider a simple example where $X$ and $Y$ are univariate and $\Z$ is two-dimensional. Specifically, in Settings~\ref{set:univariate-3}--\ref{set:univariate-4}, we consider heavy-tailed distributions.

\begin{Model}[Univariate $X$ and $Y$, and low-dimensional $\Z$]
	\label{mod:univariate}
	Let $(d_\x, d_\y, d_\z)=(1, 1, 2)$ and the sample size $n=500$.
	Each element of $\B_1 \in \bR^{d_\z \times d_\x}$, $\B_2 \in \bR^{d_\z \times d_\y}$, $\B_3 \in \bR^{d_\x \times d_\y}$, and $\Z \in \bR^{d_\z}$ are generated from $\dN(0, 1)$.
	Variables $X$ and $Y$ are generated under Settings~\ref{set:univariate-1}--\ref{set:univariate-4}, and we vary $\psi$ over the set $\{0, 0.1, 0.2, 0.3, 0.4\}$.
	For the error terms, we generate $\epsilon_\y \sim \dN(0, 1)$, and $\epsilon_\x$ is set differently.
	Let $t_3$ denote the $t$-distribution with 3 degrees of freedom.
	\begin{Setting}
		\label{set:univariate-1}
		$X = \Z \B_1 + \epsilon_\x, \quad Y = \Z \B_2 + \psi X \B_3 + \epsilon_\y, \quad \epsilon_\x \sim \dN(0, 1)$.
	\end{Setting}
	\begin{Setting}
		\label{set:univariate-2}
		$X = \Z \B_1 + \epsilon_\x, \quad Y = \Z \B_2 + \exp(\psi X \B_3) + \epsilon_\y, \quad \epsilon_\x \sim \dN(0, 1)$.
	\end{Setting}
	\begin{Setting}
		\label{set:univariate-3}
		$X = \Z \B_1 + \epsilon_\x, \quad Y = \Z \B_2 + \psi X \B_3 + \epsilon_\y, \quad \epsilon_\x \sim t_3$.
	\end{Setting}
	\begin{Setting}
		\label{set:univariate-4}
		$X = \cos(\Z \B_1) + \epsilon_\x, \quad Y = \Z \B_2 + \psi X \B_3 + \epsilon_\y, \quad \epsilon_\x \sim t_3$.
	\end{Setting}
\end{Model}

\begin{figure}[!htb]
	\centering
	\subfloat[Setting~\ref{set:univariate-1}\label{fig:sub_model_univariate_n500_s1}]{
		\includegraphics[width=0.48\linewidth]{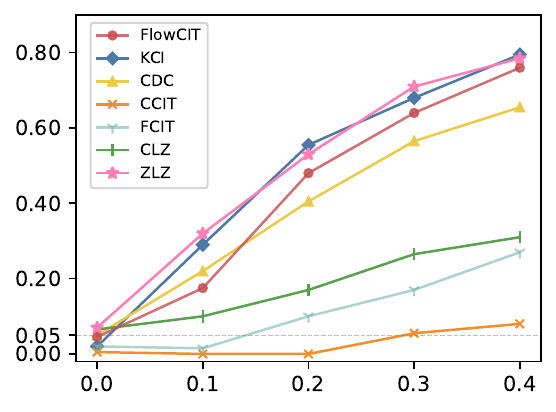}}
	\hfil
	\subfloat[Setting~\ref{set:univariate-2}\label{fig:sub_model_univariate_n500_s2}]{
		\includegraphics[width=0.48\linewidth]{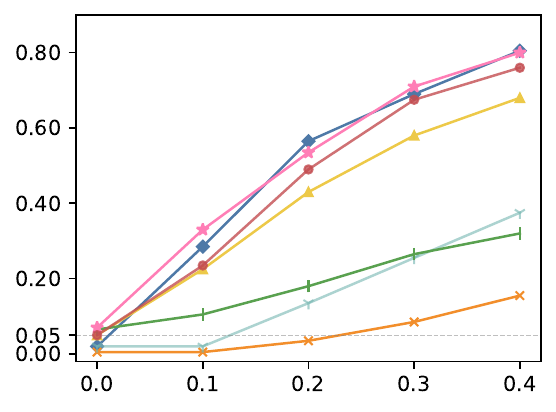}}
	\vfil
	\subfloat[Setting~\ref{set:univariate-3}\label{fig:sub_model_univariate_n500_s3}]{
		\includegraphics[width=0.48\linewidth]{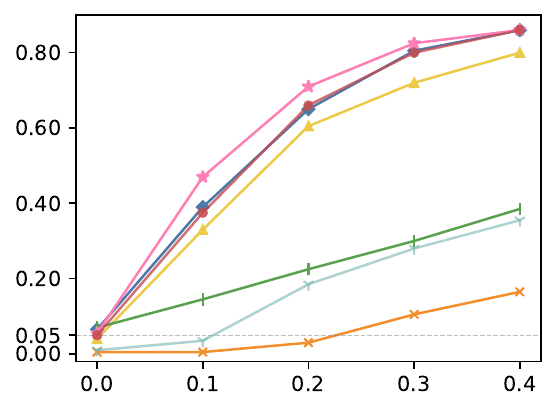}}
	\hfil
	\subfloat[Setting~\ref{set:univariate-4}\label{fig:sub_model_univariate_n500_s4}]{
		\includegraphics[width=0.48\linewidth]{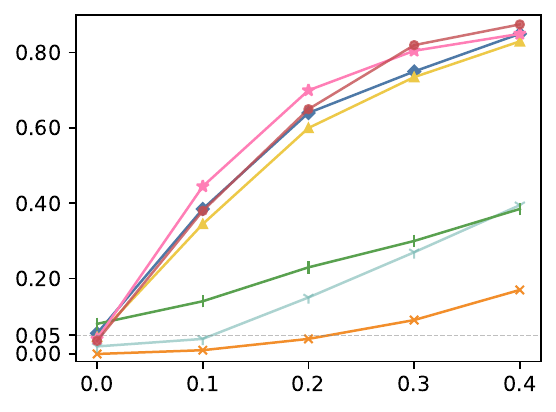}}
	\caption{
		Empirical type-I errors and statistical powers of Model~\ref{mod:univariate} with sample size $n=500$.
		The dashed gray line below refers to significance level of $0.05$.
		The $x$-axis refers to the value of $\psi$.
	}
	\label{fig:model_univariate_n500}
\end{figure}

The results of Model~\ref{mod:univariate} are shown in Figure~\ref{fig:model_univariate_n500}, where the $x$-axis represents $\psi$.
When $\psi=0$, all six methods effectively control the type-I error under $H_0$.
As $\psi$ increases, our proposed FlowCIT consistently achieves the highest statistical power under $H_1$.
The second and the third most powerful methods are KCI and CDC tests, respectively.
In contrast, CCIT is the most conservative method.

\subsection{Comparison of Two Formulations in Lemma~\ref{lemma:independence}}
\label{appendix:sec:compare_two_formula}

In this section, we refer to testing $\boldeta \indep (\bxi, \Z)$ as DC-1, and testing $\bxi \indep (\boldeta, \Z)$ as DC-2.
We demonstrate in Table~\ref{tab:dc_comparison} that testing $\boldeta \indep (\bxi, \Z)$ or $\bxi \indep (\boldeta, \Z)$ yeilds similar performance.

\begin{table}[!ht]
	\caption{Empirical type-I errors and statistical powers under different projection measures used in our FlowCIT. The results suggest that our method is robust to independence measures.}\label{tab:dc_comparison}
	\centering
	\begin{tabular}{c c | r r r r r r}
		\toprule
		& & $\psi$ & 0 & 0.1 & 0.2 & 0.3 & 0.4 \\
		\hline
		\multirow{8}{*}{Model~\ref{mod:univariate}} & \multirow{2}{*}{Setting~\ref{set:univariate-1}} & DC-1 & 0.045 & 0.175 & 0.480 & 0.640 & 0.760 \\
		&  & DC-2 & 0.050 & 0.170 & 0.485 & 0.665 & 0.735 \\
		& \multirow{2}{*}{Setting~\ref{set:univariate-2}} & DC-1 & 0.050 & 0.235 & 0.490 & 0.675 & 0.760 \\
		&  & DC-2 & 0.055 & 0.235 & 0.500 & 0.685 & 0.740 \\
		& \multirow{2}{*}{Setting~\ref{set:univariate-3}} & DC-1 & 0.050 & 0.375 & 0.660 & 0.800 & 0.860 \\
		&  & DC-2 & 0.070 & 0.380 & 0.675 & 0.810 & 0.875 \\
		& \multirow{2}{*}{Setting~\ref{set:univariate-4}} & DC-1 & 0.035 & 0.380 & 0.650 & 0.820 & 0.875 \\
		&  & DC-2 & 0.050 & 0.390 & 0.675 & 0.805 & 0.860 \\
		\midrule
		&  & $\psi$ & 0 & 0.05 & 0.1 & 0.15 & 0.2 \\
		\hline
		\multirow{8}{*}{Model~\ref{mod:model_low_low}} & \multirow{2}{*}{Setting~\ref{set:low_low-1}} & DC-1 & 0.055 & 0.210 & 0.715 & 0.955 & 0.990 \\
		&  & DC-2 & 0.065 & 0.195 & 0.710 & 0.935 & 0.980 \\
		& \multirow{2}{*}{Setting~\ref{set:low_low-2}} & DC-1 & 0.045 & 0.200 & 0.615 & 0.880 & 0.980 \\
		&  & DC-2 & 0.065 & 0.185 & 0.630 & 0.885 & 0.980 \\
		& \multirow{2}{*}{Setting~\ref{set:low_low-3}} & DC-1 & 0.055 & 0.750 & 0.965 & 1.000 & 1.000 \\
		&  & DC-2 & 0.065 & 0.750 & 0.970 & 1.000 & 1.000 \\
		& \multirow{2}{*}{Setting~\ref{set:low_low-4}} & DC-1 & 0.050 & 0.210 & 0.740 & 0.950 & 0.995 \\
		&  & DC-2 & 0.065 & 0.190 & 0.730 & 0.940 & 0.990 \\
		\midrule
		&  & $\psi$ & 0 & 0.1 & 0.2 & 0.3 & 0.4 \\
		\hline
		\multirow{8}{*}{Model~\ref{mod:model_low_high}} & \multirow{2}{*}{Setting~\ref{set:low_high-1}} & DC-1 & 0.015 & 0.640 & 0.990 & 0.995 & 1.000 \\
		&  & DC-2 & 0.025 & 0.640 & 0.990 & 0.995 & 1.000 \\
		& \multirow{2}{*}{Setting~\ref{set:low_high-2}} & DC-1 & 0.045 & 0.375 & 0.665 & 0.840 & 0.910 \\
		&  & DC-2 & 0.065 & 0.380 & 0.685 & 0.825 & 0.895 \\
		& \multirow{2}{*}{Setting~\ref{set:low_high-3}} & DC-1 & 0.050 & 0.590 & 0.980 & 1.000 & 1.000 \\
		&  & DC-2 & 0.035 & 0.575 & 0.980 & 1.000 & 1.000 \\
		& \multirow{2}{*}{Setting~\ref{set:low_high-4}} & DC-1 & 0.045 & 0.325 & 0.610 & 0.780 & 0.800 \\
		&  & DC-2 & 0.060 & 0.255 & 0.510 & 0.630 & 0.670 \\
		\midrule
		&  & $\psi$ & 0 & 0.2 & 0.4 & 0.6 & 0.8 \\
		\hline
		\multirow{8}{*}{Model~\ref{mod:model_high_high}} & \multirow{2}{*}{Setting~\ref{set:high_high-1}} & DC-1 & 0.060 & 0.380 & 0.585 & 0.750 & 0.775 \\
		&  & DC-2 & 0.070 & 0.340 & 0.605 & 0.745 & 0.800 \\
		& \multirow{2}{*}{Setting~\ref{set:high_high-2}} & DC-1 & 0.050 & 0.500 & 0.800 & 1.000 & 1.000 \\
		&  & DC-2 & 0.060 & 0.520 & 0.800 & 1.000 & 1.000 \\
		& \multirow{2}{*}{Setting~\ref{set:high_high-3}} & DC-1 & 0.045 & 0.300 & 0.750 & 0.970 & 0.995 \\
		&  & DC-2 & 0.055 & 0.290 & 0.730 & 0.950 & 1.000 \\
		& \multirow{2}{*}{Setting~\ref{set:high_high-4}} & DC-1 & 0.050 & 0.150 & 0.340 & 0.980 & 1.000 \\
		&  & DC-2 & 0.055 & 0.160 & 0.330 & 0.950 & 1.000 \\
		\bottomrule
	\end{tabular}
\end{table}

\subsection{Robustness and Comparison of Dependence Measures}
\label{appendix:sec:compare_DC_IPC}

In Section~\ref{sec:dc}, we recommend using distance correlation (DC, \citealt{szekelyMeasuring2007}) and improved projection correlation (IPC, \citealt{zhangProjective2024}) to evaluate the transformed variables due to their computational efficiency.
For both theoretical analysis and numerical studies of our proposed FlowCIT method, we adopt DC as the independence measure.
Moreover, FlowCIT exhibits robustness to the choice of independence measures.
In this section, we conduct a comparative performance analysis between DC and IPC.

According to Appendix~\ref{appendix:sec:compare_two_formula}, testing $\boldeta \indep (\bxi, \Z)$ or $\bxi \indep (\boldeta, \Z)$ yeilds similar performance. To save space, we only report the results for testing $\boldeta \indep (\bxi, \Z)$ when comparing DC and IPC.

\subsubsection{Comparison in simulations}

We conducted simulation studies aligned with the simulation setups of Models~\ref{mod:univariate}, \ref{mod:model_low_low}, \ref{mod:model_low_high}, \ref{mod:model_high_high} to evaluate the sensitivity of FlowCIT to different dependence measures. Results demonstrate that FlowCIT exhibits low sensitivity to the choice of measure, reinforcing its robustness across diverse dependency structures.

As summarized in Table~\ref{tab:dc_ipc}, the empirical type-I errors and statistical powers of FlowCIT-DC and FlowCIT-IPC are similar across all tested scenarios. This consistency highlights that the performance of FlowCIT remains stable irrespective of whether DC or IPC is employed.

\begin{table}[!htb]
\caption{Empirical type-I errors and statistical powers under different projection measures used in our FlowCIT. The results suggest that our method is robust to independence measures.}\label{tab:dc_ipc}
\centering
\begin{tabular}{c c | r r r r r r}
\toprule
 & & $\psi$ & 0 & 0.1 & 0.2 & 0.3 & 0.4 \\
\hline
\multirow{8}{*}{Model~\ref{mod:univariate}} & \multirow{2}{*}{Setting~\ref{set:univariate-1}} & DC & 0.045 & 0.175 & 0.480 & 0.640 & 0.760 \\
&  & IPC & 0.045 & 0.120 & 0.325 & 0.485 & 0.575 \\
& \multirow{2}{*}{Setting~\ref{set:univariate-2}} & DC & 0.050 & 0.235 & 0.490 & 0.675 & 0.760 \\
&  & IPC & 0.045 & 0.125 & 0.340 & 0.485 & 0.610 \\
& \multirow{2}{*}{Setting~\ref{set:univariate-3}} & DC & 0.050 & 0.375 & 0.660 & 0.800 & 0.860 \\
&  & IPC & 0.060 & 0.205 & 0.470 & 0.640 & 0.750 \\
& \multirow{2}{*}{Setting~\ref{set:univariate-4}} & DC & 0.035 & 0.380 & 0.650 & 0.820 & 0.875 \\
&  & IPC & 0.070 & 0.195 & 0.505 & 0.645 & 0.750 \\
\midrule
 &  & $\psi$ & 0 & 0.05 & 0.1 & 0.15 & 0.2 \\
\hline
\multirow{8}{*}{Model~\ref{mod:model_low_low}} & \multirow{2}{*}{Setting~\ref{set:low_low-1}} & DC & 0.055 & 0.210 & 0.715 & 0.955 & 0.990 \\
&  & IPC & 0.040 & 0.080 & 0.405 & 0.750 & 0.915 \\
& \multirow{2}{*}{Setting~\ref{set:low_low-2}} & DC & 0.045 & 0.200 & 0.615 & 0.880 & 0.980 \\
&  & IPC & 0.060 & 0.105 & 0.315 & 0.600 & 0.830 \\
& \multirow{2}{*}{Setting~\ref{set:low_low-3}} & DC & 0.055 & 0.750 & 0.965 & 1.000 & 1.000 \\
&  & IPC & 0.055 & 0.335 & 0.655 & 0.860 & 0.935 \\
& \multirow{2}{*}{Setting~\ref{set:low_low-4}} & DC & 0.050 & 0.210 & 0.740 & 0.950 & 0.995 \\
&  & IPC & 0.040 & 0.085 & 0.450 & 0.775 & 0.925 \\
\midrule
 &  & $\psi$ & 0 & 0.1 & 0.2 & 0.3 & 0.4 \\
\hline
\multirow{8}{*}{Model~\ref{mod:model_low_high}} & \multirow{2}{*}{Setting~\ref{set:low_high-1}} & DC & 0.015 & 0.640 & 0.990 & 0.995 & 1.000 \\
&  & IPC & 0.050 & 0.460 & 0.965 & 1.000 & 1.000 \\
& \multirow{2}{*}{Setting~\ref{set:low_high-2}} & DC & 0.045 & 0.375 & 0.665 & 0.840 & 0.910 \\
&  & IPC & 0.065 & 0.195 & 0.370 & 0.550 & 0.625 \\
& \multirow{2}{*}{Setting~\ref{set:low_high-3}} & DC & 0.050 & 0.590 & 0.980 & 1.000 & 1.000 \\
&  & IPC & 0.065 & 0.380 & 0.940 & 0.995 & 1.000 \\
& \multirow{2}{*}{Setting~\ref{set:low_high-4}} & DC & 0.045 & 0.325 & 0.610 & 0.780 & 0.800 \\
&  & IPC & 0.060 & 0.140 & 0.170 & 0.515 & 0.720 \\
\midrule
 &  & $\psi$ & 0 & 0.2 & 0.4 & 0.6 & 0.8 \\
\hline
\multirow{8}{*}{Model~\ref{mod:model_high_high}} & \multirow{2}{*}{Setting~\ref{set:high_high-1}} & DC & 0.060 & 0.380 & 0.585 & 0.750 & 0.775 \\
&  & IPC & 0.060 & 0.240 & 0.450 & 0.570 & 0.620 \\
& \multirow{2}{*}{Setting~\ref{set:high_high-2}} & DC & 0.050 & 0.500 & 0.800 & 1.000 & 1.000 \\
&  & IPC & 0.065 & 0.420 & 0.685 & 0.900 & 1.000 \\
& \multirow{2}{*}{Setting~\ref{set:high_high-3}} & DC & 0.045 & 0.300 & 0.750 & 0.970 & 0.995 \\
&  & IPC & 0.045 & 0.240 & 0.450 & 0.800 & 0.900 \\
& \multirow{2}{*}{Setting~\ref{set:high_high-4}} & DC & 0.050 & 0.150 & 0.340 & 0.980 & 1.000 \\
&  & IPC & 0.055 & 0.100 & 0.210 & 0.480 & 0.640 \\
\bottomrule
\end{tabular}
\end{table}

\subsubsection{Comparison in real-data analysis}

As shown in Table~\ref{tab:dc_ipc_real_data}, we also conduct a comparison between FlowCIT-DC and FlowCIT-IPC in the real-data analysis, and the resulting $p$-values are also very close.

\begin{table}[!htb]
\centering
\caption{Comparisons of $p$-values between FlowCIT-DC and FlowCIT-IPC in wine-quality dataset. Bold: $p \le 0.05$.}
\label{tab:dc_ipc_real_data}
\begin{tabular}{r c c c c}
\toprule
Methods & Supervised & Nonlinear & FlowCIT-DC & FlowCIT-IPC \\
\midrule
PCA-1 & $\xmark$ & $\xmark$ & \textbf{0.00} & \textbf{0.00} \\
PCA-4 & $\xmark$ & $\xmark$ & \textbf{0.00} & \textbf{0.00} \\
SIR-1 & $\newcheckmark$ & $\xmark$ & \textbf{0.04} & \textbf{0.02} \\
SIR-4 & $\newcheckmark$ & $\xmark$ & \textbf{0.02} & \textbf{0.02} \\
UMAP-1 & $\xmark$ & $\newcheckmark$ & \textbf{0.00} & \textbf{0.00} \\
UMAP-4 & $\xmark$ & $\newcheckmark$ & \textbf{0.00} & \textbf{0.00} \\
DDR-1 & $\newcheckmark$ & $\newcheckmark$ & \textbf{0.01} & \textbf{0.01} \\
DDR-4 & $\newcheckmark$ & $\newcheckmark$ & 0.30 & 0.48 \\
\midrule
$\calR(\X)=\X$ &  &  & 0.71 & 0.96 \\
\bottomrule
\end{tabular}
\end{table}

\subsection{Empirical Type-I Error Control}
\label{appendix:sec:typeIerror}

In the main article, we present our results using plots to effectively illustrate the behavior of each method as $\psi$ varies. Since, under $H_0$, the data points for these methods are densely clustered, we provide the empirical type-I errors in tabular form in Table~\ref{tab:empirical_size}.
As shown in Table~\ref{tab:empirical_size}, the evaluated methods achieve effective type-I error control across a majority of configurations. Mild type-I error inflation is observed in specific cases, such as CDC under the nonlinear settings of Model~\ref{mod:model_low_low} (Settings~\ref{set:low_low-2}--\ref{set:low_low-3}), though these deviations remain marginal and within acceptable thresholds.

\begin{table}[!htb]
\centering
\caption{Empirical type-I errors of Models~\ref{mod:univariate}, \ref{mod:model_low_low}, \ref{mod:model_low_high}, \ref{mod:model_high_high}.
ZLZ does not work for multivariate $\X$ and $\Y$.	
KCI and CLZ are not included in the comparisons for Models~\ref{mod:model_low_high}, \ref{mod:model_high_high} due to their high computational burden.}
\label{tab:empirical_size}
\begin{tabular}{c c | r r r r r r r}
\toprule
\multicolumn{1}{l}{} &  & FlowCIT & KCI & CDC & CCIT & FCIT & CLZ & ZLZ \\
\hline
\multirow{4}{*}{Model~\ref{mod:univariate}} & Setting~\ref{set:univariate-1} & 0.050 & 0.020 & 0.050 & 0.005 & 0.020 & 0.065 & 0.070 \\
& Setting~\ref{set:univariate-2} & 0.050 & 0.020 & 0.050 & 0.005 & 0.020 & 0.065 & 0.070 \\
& Setting~\ref{set:univariate-3} & 0.050 & 0.065 & 0.040 & 0.005 & 0.010 & 0.070 & 0.055 \\
& Setting~\ref{set:univariate-4} & 0.035 & 0.055 & 0.045 & 0.000 & 0.020 & 0.080 & 0.040 \\
\hline
\multirow{4}{*}{Model~\ref{mod:model_low_low}} & Setting~\ref{set:low_low-1} & 0.055 & 0.055 & 0.035 & 0.000 & 0.045 & 0.060 &  \\
& Setting~\ref{set:low_low-2} & 0.045 & 0.030 & 0.090 & 0.000 & 0.035 & 0.065 &  \\
& Setting~\ref{set:low_low-3} & 0.055 & 0.080 & 0.105 & 0.000 & 0.060 & 0.085 &  \\
& Setting~\ref{set:low_low-4} & 0.050 & 0.055 & 0.035 & 0.000 & 0.050 & 0.060 &  \\
\hline
\multirow{4}{*}{Model~\ref{mod:model_low_high}} & Setting~\ref{set:low_high-1} & 0.015 &  & 0.000 & 0.005 & 0.010 &  &  \\
& Setting~\ref{set:low_high-2} & 0.065 &  & 0.000 & 0.005 & 0.025 &  &  \\
& Setting~\ref{set:low_high-3} & 0.050 &  & 0.000 & 0.000 & 0.010 &  &  \\
& Setting~\ref{set:low_high-4} & 0.045 &  & 0.000 & 0.000 & 0.000 &  &  \\
\hline
\multirow{4}{*}{Model~\ref{mod:model_high_high}} & Setting~\ref{set:high_high-1} & 0.060 &  & 0.000 & 0.020 & 0.005 &  &  \\
& Setting~\ref{set:high_high-2} & 0.060 &  & 0.000 & 0.000 & 0.010 &  & \\
& Setting~\ref{set:high_high-3} & 0.045 &  & 0.000 & 0.000 & 0.000 &  & \\
& Setting~\ref{set:high_high-4} & 0.050 &  & 0.000 & 0.000 & 0.000 &  & \\
\bottomrule
\end{tabular}
\end{table}

\subsection{Implementations}
\label{sec:implement}

Among the compared methods, FlowCIT and CDC require permutation-based method, both configured with $B = 100$ permutations.
For CDC and FCIT, all other hyperparameters follow the default configurations specified in the Python package \texttt{hyppo} \citep{pandaHyppo2024}.
The KCI test is implemented using the default settings from the \texttt{R} package \texttt{CondIndTests::KCI}, while CLZ adheres to the original \texttt{R} code.
CCIT is executed with the default parameters in the \texttt{Python} package \texttt{CCIT}.
ZLZ is implemented by its original \texttt{Matlab} code.

For the FlowCIT, we utilize neural networks to learn the velocity fields.
Specifically, we design a ReLU-activated network with two hidden layers: the first layer contains $p_1$ neurons, and the second layer reduces dimensionality to $p_2 = p_1/2$ neurons.
For low-dimensional conditional independence scenarios (Models~\ref{mod:univariate} and \ref{mod:model_low_low}), we set $p_1=32$. In Model~\ref{mod:model_low_high}, we increase $p_1$ to 80, and in Model~\ref{mod:model_high_high}), we further raise $p_1$ to 600 accommodate the higher dimensionality.

\section{Proof of Lemma~\ref{lemma:independence}}
\label{sec:proof:lemma:independence}

\begin{proof}
	In this section, we present the proof for our central idea: transforming the conditional independence structure into an unconditional structure using transport maps.
	To proceed, we first prove that by construction, $\bxi \indep \Z$ and $\boldeta \indep \Z$ in the first step. We then prove the equivalance in the second step.
	
	\emph{Step 1. Prove that  $\bxi \indep \Z$ and $\boldeta \indep \Z$.}

	We show that $\bxi \indep \Z$ in the following, and $\boldeta \indep \Z$ can be proven similarly.
	
	Firstly, $\bxi$ and $\Z$ are independent if their distribution function satisfy:
	\begin{align}
	\label{eq:def_indep}
	P(\bxi, \Z) = P(\bxi) P(\Z).
	\end{align}
	Based on \eqref{eq:def_indep}, the conditional density of $(\bxi \mid \Z)$ is then given by
	\begin{align*}
	P(\bxi \mid \Z) = \frac{P(\bxi, \Z)}{P(\Z)} = P(\bxi).
	\end{align*}
	
	Thus, to prove the independence between $\bxi$ and $\Z$, it suffices to show that the law of $(\bxi \mid \Z = \z)$ coincides with the law of $\bxi$ for all $\z \in \supp(\Z)$.
	This is satisfied by:
	\begin{align*}
	\bxi = f(\X, \Z) \stackrel{d}{=} \big[ f(\X, \Z) \mid \Z = \z \big].
	\end{align*}
	
	\emph{Step 2. Prove the equivalance.}
	
	We are proving that the following three statements are equivalent:
	\begin{enumerate}
		\item $\bxi \indep \Z$, $\boldeta \indep \Z$, and $\bxi \indep \boldeta \mid \Z$.
		\item $\bxi \indep \Z$, $\boldeta \indep \Z$, and $\bxi \indep (\boldeta, \Z)$.
		\item $\bxi \indep \Z$, $\boldeta \indep \Z$, and $\boldeta \indep (\bxi, \Z)$.
	\end{enumerate}
	All these statements above can be inferred by mutual independence between $\bxi$, $\boldeta$, and $\Z$.
	Thus, it suffices to show that all these statements are equal to mutual independence between these three random vectors, which means that the joint distribution $P(\bxi, \boldeta, \Z) = P(\bxi) P(\boldeta) P(\Z)$.	
	
	Based on the first statement, it follows that:
	\begin{align*}
		P(\bxi, \boldeta, \Z) = P(\bxi, \boldeta \mid \Z) P(\Z) = P(\bxi \mid \Z) P(\boldeta \mid \Z) P(\Z) = P(\bxi) P(\boldeta) P(\Z).
	\end{align*}
	According to the second statement, it can be inferred that:
	\begin{align*}
		P(\bxi, \boldeta, \Z) = P(\bxi) P(\boldeta, \Z) = P(\bxi) P(\boldeta) P(\Z),
	\end{align*}
	and it is similar the third statement. This completes the proof of Lemma~\ref{lemma:independence}.
	
\end{proof}

\section{Proofs for the Test Statistic}
\label{sec:proofs:test}

In this section, we prove the properties of our test statistic $T_{n_2}$ as stated in Theorem~\ref{thm:asymptotics}, with detailed proof provided in Section~\ref{sec:proof:thm:asymptotics}.
The proofs of Theorem~\ref{thm:permutation_p} and Corollary~\ref{cor:multiple_split} are presented in  Section~\ref{sec:proof:thm:permutation_p}.

\subsection{Proof of Theorem~\ref{thm:asymptotics}}
\label{sec:proof:thm:asymptotics}

\begin{proof}
	Let $\U = (\bxi^\top, \Z^\top)^\top$ and $\V = \boldeta$ denote the true but not observed version while $\hat \U = (\hat \bxi^\top, \Z^\top)^\top$ and $\hat \V = \hat \boldeta$ represent the estimated counterpart.
	
	Let $a_{ij} = \| \U_i - \U_j \|$ and $b_{ij} = \| \V_i - \V_j \|$ where $i, j \in \calI_2$. Further denote the doubly centered distances:
	\begin{align*}
		A_{ij} &= a_{ij} - \bar a_{i \cdot} - \bar a_{\cdot j} + \bar a_{\cdot \cdot}, \\
		B_{ij} &= b_{ij} - \bar b_{i \cdot} - \bar b_{\cdot j} + \bar b_{\cdot \cdot},
	\end{align*}
	where $\bar a_{i \cdot}$ is the $i$-th row mean, $\bar a_{\cdot j}$ is the $j$-th column mean, and $\bar a_{\cdot \cdot}$ is the grand mean \citep{szekelyDistance2013}.
	
	We first note that $\bE (A_{ij} - \hat A_{ij})^2 = O_p(n_1^{-2\kappa_1})$, which is followed by that $a_{ij} - \hat a_{ij} = \|\U_i - \U_j\| - \|\hat \U_i - \hat \U_j\| \le \|\U_i - \hat \U_i\| + \|\U_j - \hat \U_j\|$, and the condition that $\bE_{(\X, \Z)} \| \hat \bxi - \bxi \|^2 = O_p(n_1^{-2 \kappa_1})$.
	
	 The empirical distance correlation with true $\U$ and $\V$ can be written as
	 \begin{align*}
	 	\dcov_{n_2}^2 (\U, \V) = \frac{1}{n^2} \sum_{i,j \in \calI_2} A_{ij} B_{ij}.
	 \end{align*}
	
	 For estimated $\hat \U$ and $\hat \V$, it is denoted by
	 \begin{align*}
		\dcov_{n_2}^2 (\hat \U, \hat \V) = \frac{1}{n^2} \sum_{i,j \in \calI_2} \hat A_{ij} \hat B_{ij}.
	\end{align*}	
	
	Consider the difference between these two versions:
	\begin{align*}
		& \dcov_{n_2}^2 (\U, \V) - \dcov_{n_2}^2 (\hat \U, \hat \V) \\
		&= \frac{1}{n_2^2} \sum_{i,j \in \calI_2} (A_{ij} - \hat A_{ij}) B_{ij} +
		 \frac{1}{n_2^2} \sum_{i,j \in \calI_2} A_{ij} (B_{ij} - \hat B_{ij}) +
		 \frac{1}{n_2^2} \sum_{i,j \in \calI_2} (\hat A_{ij} - A_{ij}) (B_{ij} - \hat B_{ij}) \\
		& = T_a +  T_b + T_c.
	\end{align*}
	
	We first bound the term $T_a$. Since $\hat \U_i, i \in \calI_2$ is fitted based on training data, it is independent of $\V_i, i \in \calI_2$. The expectation of $T_a$ is
	\begin{align*}
		\bE(T_a) = \frac{1}{n_2^2} \sum_{i,j \in \calI_2} \bE\{(A_{ij} - \hat A_{ij}) B_{ij}\}
		=  \frac{1}{n_2^2} \sum_{i,j \in \calI_2} \bE\{(A_{ij} - \hat A_{ij})\} \bE(B_{ij}) = 0.
	\end{align*}
	Next, the variance of $T_a$ is given by
	\begin{align*}
		\var(T_a) = \frac{1}{n_2^4} \var \Big\{ \sum_{i,j \in \calI_2}(A_{ij} - \hat A_{ij}) B_{ij} \Big\} = \frac{1}{n_2^4}\sum_{i,j,k,l \in \calI_2}  \cov \{(A_{ij} - \hat A_{ij}) B_{ij}, (A_{kl} - \hat A_{kl}) B_{kl}\}.
	\end{align*}
	For each covariance term above, if $\{i, j\} \cap \{k, l\} = \emptyset$, which means they do not share index, we have $(B_{ij}, B_{kl}) \indep \{(A_{ij} - \hat A_{ij}),  (A_{kl} - \hat A_{kl})\}$ and $\cov(B_{ij}, B_{kl}) = O(n_2^{-2})$ by the property of double-centering, and it follows that:
	\begin{align*}
		& \cov \{(A_{ij} - \hat A_{ij}) B_{ij}, (A_{kl} - \hat A_{kl}) B_{kl}\} \\
		&=  \bE\{(A_{ij} - \hat A_{ij}) B_{ij}\} \bE\{(A_{kl} - \hat A_{kl}) B_{kl}\}
		- \bE\{(A_{ij} - \hat A_{ij}) B_{ij} (A_{kl} - \hat A_{kl}) B_{kl}\} \\
		&=  \bE\{(A_{ij} - \hat A_{ij})\} \bE B_{ij} \bE\{(A_{kl} - \hat A_{kl})\} \bE B_{kl}
		- \bE\{(A_{ij} - \hat A_{ij}) (A_{kl} - \hat A_{kl})\} \bE (B_{ij} B_{kl}) \\
&= O(n_2^{-2} n_1^{-2\kappa_1}).
	\end{align*}
	Similarly, if $(i,j)$ and $(k,l)$ share one index, by the property of double-centering it follows that $\cov(B_{ij}, B_{kl}) = O(n_2^{-1})$, and we have $ \cov \{(A_{ij} - \hat A_{ij}) B_{ij}, (A_{kl} - \hat A_{kl}) B_{kl}\} = O(n_2^{-1} n_1^{-2\kappa_1})$.	
	Otherwise, if $(i, j)$ and $(k, l)$ share two indices, by the Cauchy-Schwarz inequality, we have
	\begin{align*}
		& \cov \{(A_{ij} - \hat A_{ij}) B_{ij}, (A_{kl} - \hat A_{kl}) B_{kl}\} \\
	&	\le  \sqrt{\bE(B_{ij}^2) \bE\{(A_{ij} - \hat A_{ij})^2\}\bE(B_{ij}^2) \bE\{(A_{ij} - \hat A_{ij})^2\}} = O_p(n_1^{-2\kappa_1}).
	\end{align*}
	The numbers of terms where $(i,j)$ and $(k,l)$ share no index, one index, and two indices are $O(n_2^4)$, $O(n_2^3)$, and $O(n_2^2)$. Therefore,
	\begin{align*}
		\var(T_a) = \frac{1}{n_2^4} O_p(n_2^2 \times n_1^{-2\kappa_1}) = O_p(n_2^{-2} n_1^{-2\kappa_1}).
	\end{align*}
	By Chebyshev's inequality, we have $T_a = O_p(n_2^{-1} n_1^{-\kappa_1} \log n_2)$.
	Similarly, we have $T_b = O_p(n_2^{-1} n_1^{-\kappa_2} \log n_2)$.
	
	For the term $T_c$, by Cauchy-Schwarz inequality, it follows that
	\begin{align*}
		T_c \le \sqrt{\frac{1}{n_2^2} \sum_{ij \in \calI_2}(\hat A_{ij} - A_{ij})^2 \frac{1}{n_2^2} \sum_{ij \in \calI_2} (\hat B_{ij} - B_{ij})^2  } = O_p(n_1^{-\kappa_1-\kappa_2}).
	\end{align*}
	
	Since $n_2 T_a = O_p(n_1^{-\kappa_1} \log n_2) = o_p(1)$, $n_2 T_b = o_p(1)$, and $n_2 T_c = O_p(n_2 n_1^{-\kappa_1-\kappa_2}) = o_p(1)$, the difference between $n_2 \dcov_{n_2} ^2 (\U, \V)$ and $n_2 \dcov_{n_2}^2 (\hat \U, \hat \V)$ is ignorable.
	
	Similarly, the difference between $\dcov_{n_2} ^2 (\U, \U)$ and $\dcov_{n_2} ^2 (\hat \U, \hat \U)$, as well as between $\dcov_{n_2} ^2 (\V, \V)$ and $\dcov_{n_2} ^2 (\hat \V, \hat \V)$, are also $o_p(1)$.
	Thus, by Slutsky's theorem,
	\begin{align}
		\label{eq:op1dcorr}
		n_2 T_{n_2} - n_2 \dcorr_{n_2}^2(\U, \V) = o_p(1).
	\end{align}
	
	Under $H_0$, $n_2 \dcorr_{n_2}^2(\U, \V) \stackrel{d}{\to} \sum_{j=1}^\infty \lambda_j Z_j^2$ \citep{szekelyMeasuring2007}. Together with \eqref{eq:op1dcorr}, this completes the proof for the asymptotic distribution of $T_{n_2}$ under $H_0$.
	
	Under $H_1$, it follows that $n_2 \dcorr_{n_2}(\U, \V) \stackrel{p}{\to} \infty$. We can similarly show that $\dcorr_{n_2}(\U, \V) - T_{n_2} \stackrel{p}{\to} 0$, and thus $n_2 T_{n_2} \stackrel{p}{\to} \infty$.
\end{proof}

\subsection{Proof of Theorem~\ref{thm:permutation_p} and Corollary~\ref{cor:multiple_split}}
\label{sec:proof:thm:permutation_p}

\begin{proof}
	Let $F(t)$ denote the limiting (continuous) distribution function of the test statistic $T_{n_2}$ under the null hypothesis. For a given dataset $\mathcal{D}_{n_2}$, and let $T_{n_2, b}$ denote the test statistic computed on the $b$-th permuted sample, for $b = 1, \ldots, B$. Define the empirical distribution function based on the $B$ permuted statistics as
	\begin{align*}
		\hat F_{n_2, B}(t) = \frac{1}{B} \sum_{b=1}^B \mathbf{1}\left(n_2 T_{n_2, b} \leq t\right),
	\end{align*}
	and define the permutation $p$-value as
	\begin{align*}
		p_B = 1 - \hat F_{n_2, B}(n_2 T_{n_2}),
	\end{align*}
	where $T_{n_2}$ is the observed test statistic.
	
	Since $F(t)$ is continuous and, conditional on $\mathcal{D}_{n_2}$, the $T_{n_2, b}$ are independent, it follows from the properties of permutation tests that, as $n_2 \to \infty$, the distribution of $n_2 T_{n_2, b}$ converges to $F(t)$. Therefore, by the Glivenko–Cantelli theorem, as $B, n_2 \to \infty$, $\sup_t \big| \hat F_{n_2, B}(t) - F(t) \big| \stackrel{p}{\to} 0$.
	
	Consequently, we can write
	\begin{align*}
		p_B = 1 - \hat{F}_{n_2, B}(n_2 T_{n_2}) = 1 - F(n_2 T_{n_2}) + F(n_2 T_{n_2}) - \hat{F}_{n_2, B}(n_2 T_{n_2}).
	\end{align*}
	Under the null hypothesis $H_0$, the distribution of $n_2 T_{n_2}$ converges to $F(t)$, so that $1 - F(n_2 T_{n_2}) \stackrel{d}{\to} \dU(0,1)$. Combined with the uniform convergence of $\hat{F}_{n_2, B}(t)$ to $F(t)$, it follows that $p_B \stackrel{d}{\to} \dU(0,1).$
	
	Under the alternative hypothesis $H_1$, since $n_2 T_{n_2} \stackrel{p}{\to} \infty$, we have $1 - F(n_2 T_{n_2}) \stackrel{p}{\to} 0$, and thus $p_B \stackrel{p}{\to} 0$.
	
	For Corollary~\ref{cor:multiple_split}, since the data used in each fold are disjoint, the $p$-values obtained from multiple splits are asymptotically independent, and each converges in distribution to $\dU(0,1)$ under $H_0$. Therefore, the limiting distribution of the combined $p$-value remains unchanged.

\end{proof}

\section{Proofs and Discussions for the Conditional CNFs}
\label{sec:proof:con:converge_velocity_field}


The convergence result of flow-matching stated in Assumption~\ref{con:converge_velocity_field} has been established in literature.
For instance, interested readers may refer to Theorem~5.12 of \cite{gaoConvergence2024} and Theorem~5.5 of \cite{huangConditional2024}.
Here, we use general rates $n_1^{-\kappa_1}$ and $n_1^{-\kappa_2}$ to denote. They are typically $0 < \kappa_1, \kappa_2 < 1/2$.
Based on the following convergence
\begin{align}
\label{eq:converge_from_gao}
	\bE_{\calD_{n_1}} \bE_{(t, \X, \Z)}
	\left\|\hat \v_\x(t, \X, \Z) - \v_\x(t, \X, \Z) \right\|^2 &\lesssim n_1^{-\kappa_1}.
\end{align}
Following Markov inequality, we can directly infer that,
\begin{align*}
\bE_{(t, \X, \Z)}
\left\|\hat \v_\x(t, \X, \Z) - \v_\x(t, \X, \Z) \right\|^2 \lesssim O_p(n_1^{-\kappa_1}).
\end{align*}

Based on the convergence in \eqref{eq:converge_from_gao}, the convergence of $\hat \bxi$ and $\hat \boldeta$ is followed by
\begin{align*}
& \bE_{(\X, \Z)} \|\hat \bxi - \bxi\|^2 = \bE_{(\X, \Z)} \|\int_0^1 (\hat \v_\x - \v_\x) \upd t\|^2  \\
&\le \bE_{(\X, \Z)} \int_0^1 \| (\hat \v_\x - \v_\x) \|^2 \upd t = \bE_{(t, \X, \Z)} \| \hat \v_\x - \v_\x \|^2 = O_p(n_1^{-\kappa_1}),
\end{align*}
which follows from Jensen's inequality.

\begin{singlespace}
\begin{center}
\bibliographystyle{apalike}
\bibliography{FCITarXiv0725.bib}
\end{center}
\end{singlespace}

\end{document}